\documentclass{article}
\usepackage{arxiv}
\usepackage{xcolor}

\usepackage[utf8]{inputenc} % allow utf-8 input
\usepackage[T1]{fontenc}    % use 8-bit T1 fonts
\usepackage{hyperref}       % hyperlinks
\usepackage{url}            % simple URL typesetting
\usepackage{booktabs}       % professional-quality tables
\usepackage{amsfonts}       % blackboard math symbols
\usepackage{nicefrac}       % compact symbols for 1/2, etc.
\usepackage{microtype}      % microtypography
\usepackage{lipsum}		% Can be removed after putting your text content

\usepackage{natbib}
\usepackage{doi}
\usepackage{savesym}
\usepackage{multicol}
\usepackage{multirow}
\usepackage{dsfont}

\usepackage[textsize=tiny]{todonotes}

\usepackage{mathtools}
\usepackage{graphicx}
\usepackage{times}
\usepackage{helvet}
\usepackage{courier}
\usepackage{paralist}
\usepackage{latexsym}
\usepackage{url}
\usepackage[all]{xy}
\usepackage{amsmath}
\usepackage{amssymb}
\usepackage{amsthm}
\usepackage{nccmath} % mfrac
\usepackage{wrapfig}
\usepackage{comment}
\usepackage{paralist}
\usepackage{xcolor}
\usepackage{graphicx}
\usepackage{pifont}
\usepackage{savesym}
\savesymbol{AND}
\usepackage{xspace}
\usepackage{tikz}
\usepackage{pgfplots}
\usepackage{pgf}
\usepackage{algorithm}
\usepackage{algorithmic}
\usepackage{xspace}
\usepackage{comment}
\usepackage{placeins}
\usepackage{cancel}
\usepackage{changepage}
\usepackage{bbm}
\usepackage{pgfplots}
\usepackage{filecontents}
\usepackage{placeins}
\usepackage{esvect}

\defcitealias{ch2020neural}{LS2020}
\definecolor{electricpurple}{rgb}{0.75, 0.0, 1.0}
\definecolor{darkbrown}{rgb}{0.4, 0.26, 0.13}
\definecolor{cocoabrown}{rgb}{0.82, 0.41, 0.12}
\definecolor{copper}{rgb}{0.72, 0.45, 0.2}
\definecolor{chocolate}{rgb}{0.82, 0.41, 0.12}
\usepackage{bm}

\usepackage[capitalize]{cleveref}
\if0
\usetikzlibrary{intersections}
\usetikzlibrary{arrows,calc,fit,patterns,plotmarks,shapes.geometric,shapes.misc,shapes.symbols,   shapes.arrows,   shapes.callouts,   shapes.multipart,   shapes.gates.logic.US,   shapes.gates.logic.IEC,   er,   automata,   backgrounds,   chains,   topaths,   trees,   petri,   mindmap,   matrix,   calendar,   folding, fadings,   through,   positioning,   scopes,   decorations.fractals,   decorations.shapes,   decorations.text,   decorations.pathmorphing,   decorations.pathreplacing,   decorations.footprints,   decorations.markings, shadows,circuits}
\tikzstyle{decision}=[diamond,draw]
\tikzstyle{line}=[draw]
\tikzstyle{elli}=[draw,ellipse]
\tikzstyle{arrow} = [thick]
\fi
\newcommand{\I}{\mathcal{I}}
\newcommand{\Ifeat}{\mathcal{I}^{\text{f}}}

\newcommand{\Iconv}{\mathcal{I}^{\text{cv}}}

\newcommand{\din}{d_{\text{in}}}
\newcommand{\dnet}{d_{\text{net}}}

\newcommand{\Tdgn}{\Theta^{\text{DGN}}}

\newcommand{\Tv}{{\Theta}^{\text{v}}}

\newcommand{\Tf}{\Theta^{\textrm{f}}}

\newcommand{\dc}{d_{\text{cv}}}
\newcommand{\dconv}{d_{\text{cv}}}

\newcommand{\dfc}{d_{\text{fc}}}

\newcommand{\dblock}{d_{\text{blk}}}

\newcommand{\wconv}{w_{\text{cv}}}

\newcommand{\ifout}{i_{\text{fout}}}
\newcommand{\icin}{i_{\text{cv}}}
\newcommand{\iin}{i_{\text{in}}}
\newcommand{\iout}{i_{\text{out}}}

\newcommand{\Pres}{P^{\text{res}}}
\newcommand{\Pfc}{P^{\text{fc}}}
\newcommand{\Pcnn}{P^{\text{cnn}}}

\newcommand{\eqdef}{\stackrel{\Delta}{=}}

\newcommand{\cscale}{c_{\text{scale}}}
\newcommand{\sigfc}{\sigma_{\text{fc}}}
\newcommand{\sigcnn}{\sigma_{\text{cv}}}

\newcommand{\bcnn}{\beta_{\text{cv}}}
\newcommand{\bres}{\beta_{\text{res}}}

\newtheorem{theorem}{Theorem}[section]
\newtheorem{lemma}{Lemma}[section]

\newtheorem{proposition}{Proposition}[section]

\newtheorem{assumption}{Assumption}[section]
\newtheorem{definition}{Definition}[section]

\newcommand{\J}{\mathcal{J}}
\newcommand{\ip}[1]{\langle #1\rangle}

\def\R{\mathbb{R}}

\newcounter{subequation}[equation]

\def\mathdisplay#1{%
  \ifmmode \@badmath
  \else
    $\def\@currenvir{#1}%
    \let\dspbrk@context\z@
    \let\tag\tag@in@display \SK@equationtrue %\let\label\label@in@display
    \global\let\df@label\@empty \global\let\df@tag\@empty
    \global\tag@false
    \let\mathdisplay@push\mathdisplay@@push
    \let\mathdisplay@pop\mathdisplay@@pop
    \if@fleqn
      \edef\restore@hfuzz{\hfuzz\the\hfuzz\relax}%
      \hfuzz\maxdimen
      \setbox\z@\hbox to\displaywidth\bgroup
        \let\split@warning\relax \restore@hfuzz
        \everymath\@emptytoks \m@th $\displaystyle
    \fi
}

\usepackage{tabularx}
\usepackage{graphicx}
\usepackage{amsmath}
\title{Explicitising the implicit interpretability of Deep Neural Networks via Duality}

\author{ Chandrashekar Lakshminarayanan\\
Indian Institute of Technology Madras\\
\texttt{chandrashekar@cse.iitm.ac.in}\\
\And
Amit Vikram Singh\\
\texttt{amitkvikram@gmail.com}\\
\And
Arun Rajkumar\\
Indian Institute of Technology Madras\\
\texttt{arunr@cse.iitm.ac.in}\\
}

\renewcommand{\shorttitle}{\textit{Explicitising the implicit interpretability of DNNs}}

\hypersetup{
pdftitle={Explicitising the implicit interpretability of DNNs},
pdfsubject={cs-ml-dl},
pdfauthor={Chandrashekar Lakshminarayanan, Amit Vikram Singh},
pdfkeywords={First keyword, Second keyword, More},
}

\begin{document}
\maketitle

\begin{abstract}
Recent work by \cite{npk} provided a dual view for fully connected deep neural networks (DNNs) with rectified linear units (ReLU). It was shown that (i) the information in the gates is analytically characterised by a kernel called the neural path kernel (NPK) and (ii) most critical information is learnt in the gates, in that, given the learnt gates, the weights can be retrained from scratch without significant loss in performance.  Using the dual view, in this paper, we rethink the conventional interpretations of DNNs thereby explicitsing the implicit interpretability of DNNs. Towards this, we first show new theoretical properties namely rotational invariance and ensemble structure of the NPK in the presence of convolutional layers and skip connections respectively. Our theory leads to two surprising empirical results that challenge conventional wisdom: (i) the weights can be trained even with a constant $\mathbf{1}$ input, (ii) the gating masks can be shuffled, without any significant loss in performance. These results motivate a novel class of networks which we call deep linearly gated networks (DLGNs). DLGNs using the phenomenon of \emph{dual lifting} pave way to more direct and simpler interpretation of DNNs as opposed to conventional interpretations. We show via extensive experiments on CIFAR-10 and CIFAR-100 that these DLGNs lead to much better interpretability-accuracy tradeoff.
 \end{abstract}

\section{Introduction}\label{sec:intro}
Despite their success deep neural networks (DNNs) are still largely considered as black boxes.  The main issue is that in each layer of a DNN, the linear computation, i.e., multiplication by the weight matrix  and the non-linear activations are entangled. The conventional interpretation  is that such entanglement is the key to success of DNNs, in that, it allows DNNs to learn sophisticated structures in a layer-by-layer manner. In this work, we rethink this conventional interpretation using the recently developed dual view of DNNs \cite{npk}.

Duality for DNNs with rectified linear units (ReLUs) attempts to understand DNNs via paths going from input to output instead of the conventional layer-by-layer view. Thanks to the gating property (i.e., `on/off' states) of the ReLUs, the output can be expressed as a summation of contribution of the paths. The information in the gates is encoded in the so called \emph{neural path feature} vector and the information in the weights are encoded in the \emph{neural path value} vector, both of which are vectors in the dimension of the total number of paths. It follows then that the output is the inner product of the neural path feature and neural path value - this separates the gates from the weights analytically. 

\begin{figure}
    \centering
    \begin{minipage}{0.5\columnwidth}
    \centering
        \begin{tikzpicture}
\node []  (fntext)at (5-3,1+0.25) {\tiny{Deep Gated Network}};
%Feature Network
\node [draw,
	minimum width=2cm,
	minimum height=0.625cm,
]  (fnbox)at (5-3,0.375+0.25) {};
\node []  (fntext)at (5-3,0.5+0.25) {\tiny{Gating Network}};

\node []  (fntext)at (5-3,0.25+0.25) {\tiny{DNN with ReLUs}};

%Feature Network Input
\node (fin) [left of=fnbox,node distance=1.25cm, coordinate] {};
\node[left=-1pt] at (fin.west){\tiny{Input}};
\draw[-stealth] (fin.center) -- (fnbox.west);

%Value Network

\node [draw,
	minimum width=2cm,
	minimum height=0.625cm,
]  (vnbox)at (5-3,-0.875) {};
\node []  (fntext)at (5-3,-0.75) {\tiny{Weight Network}};
\node []  (vntext)at (5-3,-1) {\tiny{DNN with GaLUs}};

%Value Network Input
\node (vin) [left of=vnbox,node distance=1.25cm, coordinate] {};
\node[left=-1pt] at (vin.west){\tiny{Input}};
\draw[-stealth] (vin.center) -- (vnbox.west);

%Feature Network Output
\node (vout) [right of=vnbox,node distance=1.25cm, coordinate] {};
\node[right=-1pt] at (vout.west){\tiny{Output}};
\draw[-stealth]  (vnbox.east)--(vout.center);

\draw[-stealth]  (fnbox.south)--(vnbox.north);

\node []  (gates)at (5.75-3,-0.125) {\tiny{Gating Signal}};

\end{tikzpicture}
    \end{minipage}
    \caption{Deep Gated Network}
    \label{fig:dgn}
\end{figure}
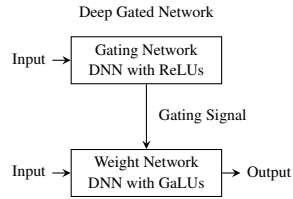

To understand the role of the gates, a deep gated network (DGN) (see \Cref{fig:dgn}) was used to separate the gates from the weights. 
Using the DGN (as an experimental setup), two important insights were provided:  (i) learning in the gates is the most crucial, in that, by using learnt gates in the gating network (by pre-training it), and retraining the weight network from scratch, the DGN can match the performance of the DNN, and (ii) in the limit of  infinite width, learning the weights with fixed gates is equal to a kernel method with the neural path kernel (NPK)\footnote{\cite{npk} showed that in a DGN the NPK is equal to a constant times the \emph{neural tangent kernel} (NTK). The fact that infinite width DNNs are equivalent to the NTK was shown in  \citep{ntk,arora2019exact,cao2019generalization}.} which is the kernel associated with the neural path features. 

While \cite{npk} provide an analytical expression for the NPK, the structural properties of the NPK were not entirely explored in their work. As we will see later, understanding these structural properties perhaps surprisingly leads to a fundamental rethinking of the conventional interpretation of DNNs. We establish this in this paper via novel theoretical and empirical contributions as listed below.

\begin{figure*}[!t]
\centering
\begin{minipage}{0.75\columnwidth}
\resizebox{1.0\columnwidth}{!}{
\begin{tabular}{cc|c|cccccccc|c|c|c@{}|}\cline{3-13}
&&\resizebox{150pt}{!} {\Huge{Input}}& \resizebox{200pt}{!}{\shortstack{\Huge{Layer 1}\\\Huge{Filter 1}}}& \resizebox{200pt}{!}{\shortstack{\Huge{Layer 1}\\\Huge{Filter 2}}}&
\resizebox{200pt}{!}{\shortstack{\Huge{Layer 2}\\\Huge{Filter 1}}}& \resizebox{200pt}{!}{\shortstack{\Huge{Layer 2}\\\Huge{Filter 2}}}&
\resizebox{200pt}{!}{\shortstack{\Huge{Layer 3}\\\Huge{Filter 1}}}& \resizebox{200pt}{!}{\shortstack{\Huge{Layer 3}\\\Huge{Filter 2}}}&
\resizebox{200pt}{!}{\shortstack{\Huge{Layer 4}\\\Huge{Filter 1}}}& \resizebox{200pt}{!}{\shortstack{\Huge{Layer 4}\\\Huge{Filter 2}}}&
\resizebox{150pt}{!}{\shortstack{\Huge{Test}\\\Huge{ Acc.}}}&\resizebox{380pt}{!}{\shortstack{\Huge{Visual}\\\Huge{Interpretation}}}\\
\cline{2-13}
\resizebox{150pt}{!} {\shortstack{\quad~{}}}&
\multicolumn{1}{|c|}{\resizebox{400pt}{!} {\shortstack{\Huge{Gating Network}\\\Huge \\\mbox{}\\\mbox{}\\\mbox{}\\\mbox{}\\\mbox{}\\\mbox{}\\\mbox{}}}}&
\includegraphics{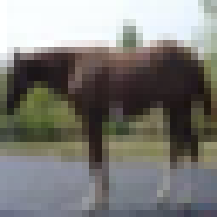}&
\includegraphics{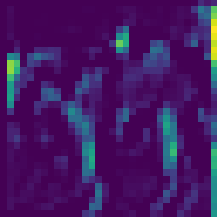}&
\includegraphics{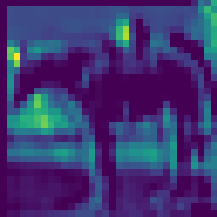}&
\includegraphics{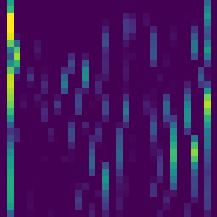}&
\includegraphics{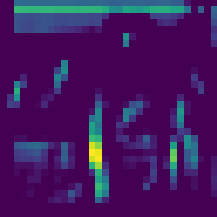}&
\includegraphics{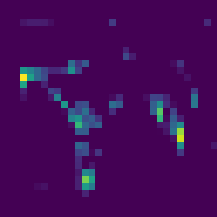}&
\includegraphics{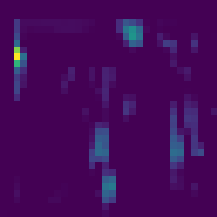}&
\includegraphics{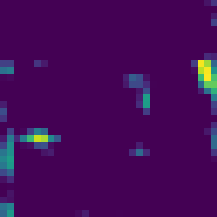}&
\includegraphics{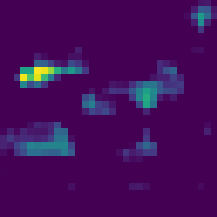}&
{\resizebox{200pt}{!}{\shortstack{80.4\tiny{$\pm$ 0.3}\\\mbox{}\\\mbox{}\\\mbox{}\\\mbox{}}} }&
\resizebox{100pt}{!}{\shortstack{\mbox{}\mbox{}\ding{51}\\\mbox{}\mbox{}\mbox{}\mbox{}}}\\\hline
\multicolumn{1}{|c}{\resizebox{150pt}{!} {\shortstack{\Huge{DGN-1}\\\mbox{}\\\mbox{}\\\mbox{}\\\mbox{}\\\mbox{}\\\mbox{}}}}&
\multicolumn{1}{|c|}{\resizebox{400pt}{!} {\shortstack{\Huge{Weight Network}\\\Huge \\\mbox{}\\\mbox{}\\\mbox{}\\\mbox{}\\\mbox{}\\\mbox{}\\\mbox{}}}}&
\includegraphics{images_neurips_2021/horse.png}&
\includegraphics{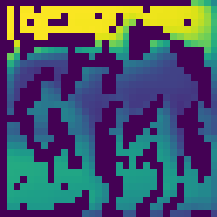}&
\includegraphics{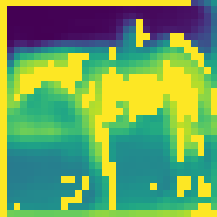}&
\includegraphics{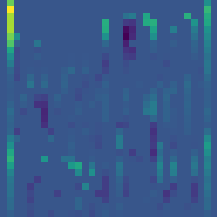}&
\includegraphics{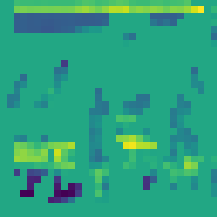}&
\includegraphics{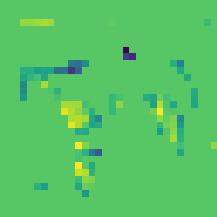}&
\includegraphics{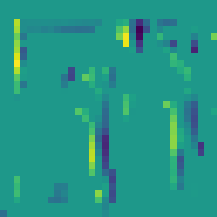}&
\includegraphics{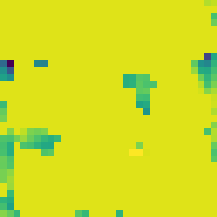}&
\includegraphics{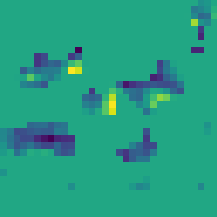}&
{\resizebox{200pt}{!}{\shortstack{79.0\tiny{$\pm$ 0.2}\\\mbox{}\\\mbox{}\\\mbox{}\\\mbox{}}} }&
\resizebox{100pt}{!}{\shortstack{\mbox{}\mbox{}\ding{53}\\\mbox{}\mbox{}\mbox{}\mbox{}}}\\\hline
\multicolumn{1}{|c|}{\resizebox{150pt}{!} {\shortstack{\Huge{DGN-2}\\\mbox{}\\\mbox{}\\\mbox{}\\\mbox{}\\\mbox{}\\\mbox{}}}}&
\multicolumn{1}{c|}{\resizebox{400pt}{!} {\shortstack{\Huge{Weight Network}\\\Huge{$\mathbf{1}$ input} \\\mbox{}\\\mbox{}\\\mbox{}\\\mbox{}\\\mbox{}\\\mbox{}\\\mbox{}}}}&
\includegraphics{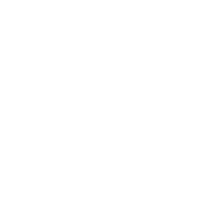}&
\includegraphics{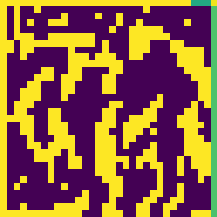}&
\includegraphics{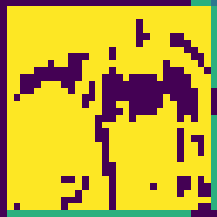}&
\includegraphics{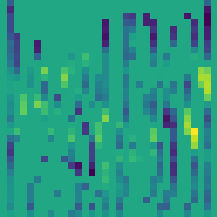}&
\includegraphics{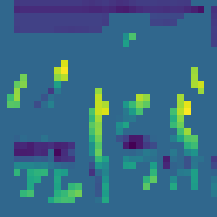}&
\includegraphics{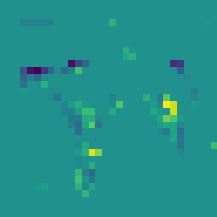}&
\includegraphics{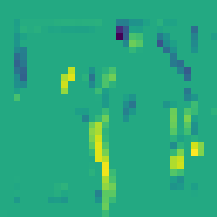}&
\includegraphics{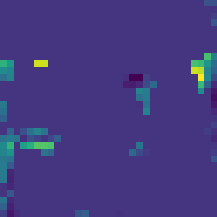}&
\includegraphics{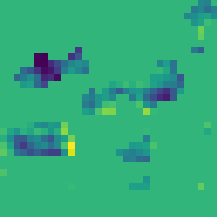}&
{\resizebox{200pt}{!}{\shortstack{79.5\tiny{$\pm$ 0.2} \\\mbox{}\\\mbox{}\\\mbox{}\\\mbox{}}} }&
\resizebox{100pt}{!}{\shortstack{\mbox{}\mbox{}\ding{53}\\\mbox{}\mbox{}\mbox{}\mbox{}}}\\\hline
\multicolumn{1}{|c|}{\resizebox{150pt}{!} {\shortstack{\Huge{DGN-3}\\\mbox{}\\\mbox{}\\\mbox{}\\\mbox{}\\\mbox{}\\\mbox{}}}}&
\multicolumn{1}{c|}{\resizebox{400pt}{!} {\shortstack{\Huge{Weight Network}\\\quad\\\mbox{}\\\Huge{$\mathbf{1}$ input + }\\\Huge{Layer Shuffling} \mbox{}\\\mbox{}\\\mbox{}\\\mbox{}\\\mbox{}}}}&
\includegraphics{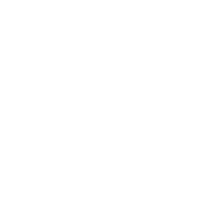}&
\includegraphics{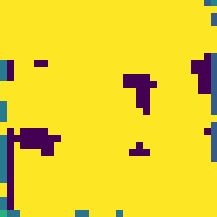}&
\includegraphics{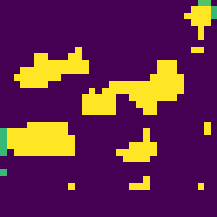}&
\includegraphics{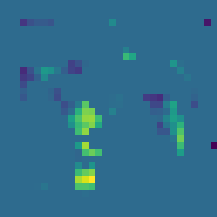}&
\includegraphics{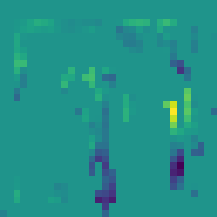}&
\includegraphics{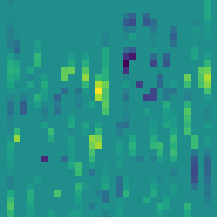}&
\includegraphics{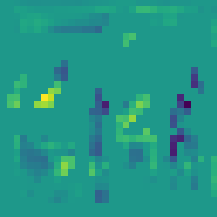}&
\includegraphics{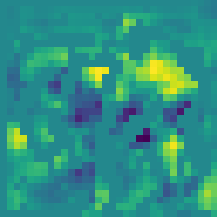}&
\includegraphics{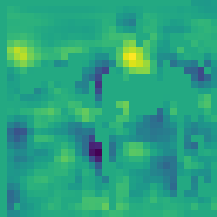}&
{\resizebox{200pt}{!}{\shortstack{79.3\tiny{$\pm$0.3} \\\mbox{}\\\mbox{}\\\mbox{}\\\mbox{}}} }&
\resizebox{100pt}{!}{\shortstack{\mbox{}\mbox{}\ding{53}\\\mbox{}\mbox{}\mbox{}\mbox{}}}
\\\bottomrule
\end{tabular}
}
\end{minipage}
\caption{\small{Shows the hidden layer outputs of $3$ different convolutional DGNs namely DGN-1/2/3 (each with 4 hidden layers and $128$ filters per layer). Each row has the input image to the network followed by the output of first $2$ filters in each of the $4$ hidden layers.  All $3$ DGNs have the same gating network which is shown in the top row. The 3 DGNs differ in the weight network as follows: (i) in DGN-1 in the second row, the weight network is provided with the input image and the layers are not shuffled, (ii) in DGN-2 in the third row, the weight network is provided with $\mathbf{1}$ as input and the layers are not shuffled, and (iii)  in DGN-3 in the final row, the weight network is provided with $\mathbf{1}$ as input and the layers are shuffled. In the last two rows, the entry in the input image column `appears' blank because it is the image of the $\mathbf{1}$ input to the weight network.}}
\label{fig:horse}
\end{figure*}

\begin{table*}[ht]
\centering
\begin{tabular}{|c|c|c|c|}\cline{2-4}
\multicolumn{1}{c|}{}&\textbf{DNN} (Standard)&\textbf{DGN} (\citep{npk})&\textbf{DLGN} \color{blue}{(this paper)}\\\hline
\multicolumn{1}{|c|}{\multirow{2}{*}{\textbf{Gating Network}}}&\multirow{3}{*}{\shortstack{Not separate\\ Non-linearly entangled\\ \color{red}{Non-interpretable}}}
&\multirow{2}{*}{\shortstack{Non-linear and hence\\ \color{red}{Non-interpretable}}}&\multirow{2}{*}{\shortstack{\color{green}Linear and hence \\{\color{green}{interpretable}} \color{blue}{(this paper)}}}\\
& 					&  		&\\ \cline{1-1} \cline{3-4}
\multicolumn{1}{|c|}{\textbf{Weight Network}}&&\multicolumn{2}{c|}{{\color{brown}{Non-linear but unncessary for interpretability}} \color{blue}{(this paper)}}\\\hline
\end{tabular}
\caption{Upshot of the contributions in this paper}
\label{tb:insights}
\end{table*}
\textbf{Theoretical Contributions}:  We present (\Cref{sec:theory}) an unnoticed insight in prior work on fully connected networks that the NPK is a \emph{product kernel} and is invariant to layer permutations. We present new results to show that (i) the NPK is \emph{rotationally invariant} for convolutional networks with global average pooling, and (ii) the NPK is an \emph{ensemble of many kernels} in the presence of skip connections.

\textbf{Key Empirical Contributions:}
These theoretical results inspire the following empirical questions which lead to conventionally unexpected results.

\textit{Constant Input:} We show that providing a constant $\mathbf{1}$ as input to the weight network of the DGN does not degrade performance. In the absence of `useful' input,  the conventional view that starting from the input, sophisticated features are learnt in a layer-by-layer manner fails to explain why the network still learns only with a constant $\mathbf{1}$ as input. 

The invariance structure of the NPK allows us to explicitly break the layer-by-layer structure as follows.

\textit{Layer Shuffling:} We show that providing constant input together with permuting the layers (and then applying them as external masks) still does not degrade performance. We infer from our results that looking for `visually' interpretable representations in the hidden layer outputs of the weights network is not meaningful and is not even an inherent property of the learnt weight network (see \Cref{fig:horse}) .

\textbf{Modeling Contribution:}
%The above experiments clearly show that non-linear weight network is not the source of interpretability. Thus, it is natural to look for interpretability in the gating network. However, current DGNs have non-linearity in the gating network making them non interpretable. To overcome this, we propose a novel architecture which we call the Deep Linearly Gated Networks (DLGN).
The insights we derive from the above surprising experiments are summarized in \Cref{tb:insights}. In short, we conclude that DGNs not only provide a way to separate the weights from the gates, but more importantly also tell us that one must \emph{not} look for \emph{interpretability} in the weight network. 

\textit{Deep Linearly Gated Networks (DLGN):}  Once we argue that the weight network has no interpretability value, we turn to the gating network. A standard DGN's gating network has non-linear activations. We hypothesize that this is not essential. Towards this, we propose a novel architecture called Deep Linearly Gated Network which  entirely eliminates the ReLUs (non-linearity) from the gating network thus making it a deep linear network. Our experiments (on CIFAR-10 and CIFAR-100) show that the DLGNs have much better interpretability-accuracy tradeoff than what a conventional route would suggest. We attribute and explain this using a phenomenon called \emph{dual lifting} in DGN/DLGNs.

Our results show that the only useful source of interpretability in deep networks is in the linear gating network. This rethinking allows us to argue for the case that the deep learning community's attempts at developing inbuilt algorithmic interpretability models must be via deep linear gating networks (DLGN).

\textbf{Related Works.} We now discuss prior works related most to our work and defer a more elaborate discussion to the appendix. \\
\textbf{Kernels:} Several works have examined theoretically as well as empirically two important kernels associated with a DNN namely its NTK based on the correlation of the gradients and the conjugate kernel based on the correlation of the outputs \citep{spectra,laplace,belkin,genntk,disentangling,ntk,arora2019exact,convgp,fcgp,lee2020finite}. In contrast, the NPK is based on the correlation of the gates. We do not build pure-kernel method with NPK, but use it as an aid to rethinking the conventional `hidden layer' interpertation of finite width DNNs.\\  %It was shown in [\citenum{li2019enhanced}], that  prediction using CNTK with GAP is equivalent to prediction using CNTK without GAP but with full translation data augmentation with wrap-around at the boundary. This is related to \Cref{th:mainconv}. It was shown in [\citenum{veit2016residual}] that residual networks behave like ensemble of shallow networks. This is related \Cref{th:mainres}.
%\textbf{Random Labels.} In [\citenum{randlabel}], both positive and negative effects on downstream training performance due to upstream training with random labels was studied. The question of why the test performance degrades due to upstream training with random labels was left open, which we addressed in our paper. 
\begin{comment}
$\bullet$ \textbf{Finite vs Infinite Width.} \cite{finitevsinfinite} perform an extensive comparison of finite versus infinite width DNNs. An aspect that is absent in their work, but present in the dual view is the disentanglement of gates and weights, and the fact that the learning in gates is crucial for finite width network to outperform infinite width DNNs. In our paper, we make use of theory developed for infinite width DNNs to provide empirical insights into inner workings of finite width networks.\\
\end{comment}
\textbf{Capacity:} Our experiments on destruction of layers, and providing constant $\mathbf{1}$ input are direct consequences of the insights from dual view theory. These are not explained by mere capacity based studies showing  DNNs are powerful to fit even random labelling of datasets \citep{ben}.
\textbf{Interpretability:} \cite{rudinstop} provides  several compelling arguments for why it is important to build models that are interpretable by design as opposed to seeking \emph{post-hoc} explanations for decisions of non-interpretable models. \cite{rudininterpretable} discusses the grand challenges in interpretablity. The DLGN in our paper resonates with the following aspect discussed in \cite{rudinstop,rudininterpretable}, i.e., most times the simpler models that are used to explain the decision of the more complicated non-interpretable models do not mimic the computations, thereby the explanations are not faithful which is an issue. The DLGN in our paper does not suffer from this issue because, the DLGN is obtained by rearranging the computations of a DNN with ReLUs in a mathematically principled manner.

\section{Neural Path Kernel and Dual View}\label{sec:prelim}
In this section, we first describe the recently developed dual path view of DNNs with ReLUs \citep{npk}. Among other things, the dual view leads to an understanding of (infinite width) DNNs as kernel methods with a novel kernel called the Neural Path Kernel (NPK). The NPK is related to the popular Neural Tangent Kernel (NTK). For a pair of input examples $x,x'$  and network weights $\Theta\in\R^{\dnet}$, the NTK is defined as follows:
%\textbf{NTK.}  An important kernel associated with a DNN is its \emph{neural tangent kernel} (NTK), which, for a pair of input examples $x,x'\in\R^{\din}$, and network weights $\Theta\in\R^{\dnet}$, is given by:

{\centering  $\text{NTK}(x,x')\quad = \quad \ip{\nabla_{\Theta}\hat{y}(x), \nabla_{\Theta}\hat{y}(x')}$,\quad\text{where}\par}
$\hat{y}_\Theta(\cdot)\in\R$ is the DNN output. Prior works \citep{ntk,arora2019exact,cao2019generalization} have shown that, as the width of the DNN goes to infinity, the NTK matrix converges to a limiting deterministic matrix $\text{NTK}_{\infty}$, and training an infinitely wide DNN is equivalent to a kernel method with $\text{NTK}_{\infty}$.

%\subsection{Dual View For DNNs with ReLUs: Characterising the role of gates}
The dual view gives a more fine grained understanding of a fully connected DNN with `$d$' layers and `$w$' hidden units in each layer. Specifically, in the dual view, the computations are broken down path-by-path. The input and the gates (in each path) are encoded in a \emph{neural path feature vector} - $\phi_{\Theta}(x) \in \R^P$ and the weights (in each path) are encoded in a \emph{neural path value vector} - $v_{\Theta}(x) \in \R^P$, where $P$ is the total number of paths.   
The output of the DNN was shown to be the inner product of these two vectors: 
\begin{align}\label{eq:inner}
\hat{y}_{\Theta}(x)=\ip{\phi_{\Theta}(x),v_{\Theta}}=\sum_{p=1}^P  \phi_{\Theta}(x,p) v_{\Theta}(p)
\end{align}
Furthermore, using the dual view, a novel kernel called NPK was defined as follows:
$$\text{NPK}_{\Theta}(x,x')\eqdef \ip{\phi_{\Theta}(x),\phi_{\Theta}(x')}= \ip{x,x'}{\textbf{overlap}_{\Theta}(x,x')} $$
where $\textbf{overlap}_{\Theta}(x,x')$ is equal to the total number of paths that get activated for both inputs $x$ and $x'$. 

The dual view led \cite{npk} to the question of seeking the \emph{real source of learning} in DNNs. To understand this, they propose the  \textbf{Deep Gated Network (DGN)} which is a setup to separate the gates from the weights. Consider a DNN with ReLUs with weights $\Theta\in\R^{\dnet}$. The DGN \emph{corresponding} to this DNN (\Cref{fig:dgn}) has two networks of \emph{identical architecture} (to the DNN) namely the gating network and the weight network with distinct weights $\Tf\in\R^{\dnet}$ and $\Tv\in\R^{\dnet}$.  
%The main difference between the feature and value networks is in the activations. 
The gating network has ReLUs which turn `on/off' based on their pre-activation signals, and the weight network has gated linear units (GaLUs) \citep{sss,npk}, which multiply their respective pre-activation inputs by the external gating signals provided by the gating network.  Since both the networks have identical architecture, the ReLUs and GaLUs in the respective networks have a one-to-one correspondence.  Gating network realises $\phi_{\Tf}(x)$ by turning `on/off' the corresponding GaLUs in the weight network. The weight network realises $v_{\Tv}$ and computes the output $\hat{y}_{\text{DGN}}(x)=\ip{\phi_{\Tf}(x),v_{\Tv}}$.

The main result of \cite{npk} shows that under mild conditions, for a fully connected DGN, the NTK and NPK are related by a constant factor. Specifically, in the limit of infinite width, assuming all the weights are initialized from the set $\{\sigma, -\sigma\}$ with equal probability, they show the following:
\begin{align*}
\text{NTK}(x,x')\,\,\rightarrow\,\, &d \cdot \sigma^{2(d-1)} \cdot \text{NPK}(x,x'), \quad\text{as}\,\, w\rightarrow \infty \\
\end{align*}
Using extensive experiments in the DGN setup that separates the gates from the weights, \cite{npk} were able to demonstrate that the real source of learning is in the gates and that learning the gates is key for good performance.  

\section{Theoretical Results}\label{sec:theory}
In this section, we give novel theoretical results concerning the Neural Path Kernel. These results will help us understand DGNs better and will lead us to devise novel experiments with conventionally unexpected results. 
Our first result is based on a simple but very significant observation about the structure of the Neural Path Kernel. Recall that the NPK depends on $\textbf{overlap}(x,x')$. 
We show that this term can in fact be written in a product form and so the NPK kernel is infact a product kernel.  
\begin{theorem}{(\textbf{NPK is a Product Kernel})}
\label{th:fc} Let $G_l(x)\in[0,1]^w$ denote the gates in layer $l\in\{1,\ldots,d-1\}$ for input $x\in\R^{\din}$. Then 
$$\textbf{overlap}(x,x')=\Pi_{l=1}^{(d-1)}\ip{G_l(x),G_l(x')} $$
Thus the NPK for a fully connected DGN is given by
\begin{align*}
\text{NPK}(x,x') =  \ip{x,x'}\cdot \Pi_{l=1}^{(d-1)}\ip{G_l(x),G_l(x')}
\end{align*}
\end{theorem}
%Furthermore, under the same assumptions as \cite{npk}, the following holds: ($\sigma=\frac{\cscale}{\sqrt{w}}$) as $w\rightarrow \infty $, we have for fully connected DGN:
%\begin{align*}
%\text{NTK}(x,x')  =  d \cdot \cscale^{2(d-1)} \cdot \left(\ip{ x,x'} \cdot \Pi_{l=1}^{d-1} \frac{\ip{G_l(x),G_l(x')}}w\right),
%\end{align*}
%\end{theorem} 

\Cref{th:fc} provides the most simplest kernel expression that characterises the information in the gates. The product structure, as we will see later, is key to devise experiments that break the layer by layer structure of DGNs.  

While only fully connected DGNs were analyzed in \cite{npk}, we go further and derive the structure of NPK for two important architectures - ConvNets with global-average-pooling and ResNet with skip connections.  The next two theorems make this precise. In what follows, let the circular rotation of vector $x\in\R^{\din}$ by `$r$' co-ordinates be defined as $rot(x,r)(i)=x(i+ r)$, if $i+r \leq \din$ and $rot(x,r)(i)=x(i+ r-\din)$ if $i+r > \din$.
 
 \begin{theorem}{ (\textbf{NPK for ConvNets is rotationally invariant})}\label{th:conv} The Neural Path Kernel for a Convolutional Neural Network is given as follows:
 \begin{align*}
\text{NPK}^{\texttt{CONV}}(x,x') = \sum_{r=0}^{\din-1} \ip{x,\Lambda rot(x',r)}
\end{align*}
where $\Lambda = {\textbf{overlap}(\cdot, x,rot(x',r))}$.
 \end{theorem}
Here $\Lambda$ is a diagonal matrix whose diagonal entries correspond to the total number of common paths starting from input node $i$ active for both inputs $x$ and $x'$. Note that in the fully connected case, $\textbf{overlap}(i,x,x')$ is identical for all $i=1,\ldots,\din$, and hence $\ip{x,\Lambda x'}$  becomes $\ip{x,x'}\cdot \textbf{overlap}(x,x')$. This in general will not be the case for ConvNets.

The right hand side of the expression for $\text{NPK}^{\texttt{CONV}}$ can further be written as $\sum_{r=0}^{\din-1}\sum_{i=1}^{\din} x(i) rot(x',r)(i)\textbf{overlap}(i,x,rot(x',r))$, where the inner `$\Sigma$' is the inner product between $x$ and $rot(x',r)$ weighted by $\textbf{overlap}$ and the outer `$\Sigma$' covers all possible rotations, which in addition to the fact that all the variables internal to the network rotate as the input rotates, results in the rotational invariance.  That said, rotational invariance is not a new observation; it was shown by \cite{li2019enhanced} the NTK for ConvNets with global-average-pooling are rotationally invariant. Importance of \Cref{th:conv} lies in the fact that it gives a fine grained expression for the kernel by explicitising the gates, a fact which will be used critically in the experiments.

We next consider a ResNet with `$(b+2)$' blocks and `$b$' skip connections between the blocks. Each block is a fully connected (FC) network of depth `$\dblock$' and width `$w$'. There are $2^b$ many sub-FCNs within this ResNet.
Note that the blocks being fully connected is for expository purposes, and the result continue to hold for any kind of block.

Let $2^{[b]}$ denote the power set of $[b]$ and let $\J\in 2^{[b]}$ denote any subset of $[b]$. Define the`$\J^{th}$' sub-FCN of the ResNet to be the fully connected network obtained by (i) including  $\text{block}_{j},\forall j\in \J$  and (ii) ignoring $\text{block}_{j},\forall j\notin \J$. In \Cref{th:res} below, $\text{NPK}^{\texttt{FC}}_{\J}$ is the NPK of the $\J^{th}$ FCN.

\begin{theorem}{\textbf{(NPK for ResNet is an Ensemble)}}\label{th:res} 
The NPK structure for Res-Net is given by
\begin{align*}
\text{NPK}^{\texttt{RES}} =  \sum_{\J\in 2^{[b]}} \text{NPK}^{\texttt{FC}}_{\J}, \,\, 
\end{align*}
\end{theorem}
To the best of our knowledge, this is the first theoretical result to show that ResNets have an ensemble structure, where  each kernel in the ensemble, i.e., $\text{NPK}^{\texttt{FC}}_{\J}$ corresponds to one of the $2^b$ sub-architectures. We will see in the next section how the ensemble property leads to surprising empirical consequences.

\textbf{Remark:} We also derive that in limit of infinite width, the NTKs,   $\text{NTK}^{\texttt{CONV}}$ and $\text{NTK}^{\texttt{RES}}$ is equal to a constant times $\text{NPK}^{\texttt{CONV}}$ and $\text{NPK}^{\texttt{RES}}$. We defer the formal results to the appendix as these are not the main focus of this paper.

\begin{table*}[t]
\centering
\begin{tabular}{|c|c|c|c|c|c|c|}\cline{4-7}
\multicolumn{3}{c}{}&\multicolumn{2}{c|}{\texttt{WITHOUT SHUFFLING}} &\multicolumn{2}{c|}{\texttt{WITH SHUFFLING}} \\\hline
Architecture & Dataset & DNN & DGN & DGN-\texttt{ALLONES} & DGN & DGN-\texttt{ALLONES}\\\hline
 FC4 &MNIST      &98.5\tiny{$\pm$0.1}       &98.6\tiny{$\pm$0.1}           &98.6\tiny{$\pm$0.1} &98.6\tiny{$\pm$0.1}           &98.6\tiny{$\pm$0.1}       \\   
 CONV4&CIFAR-10   &80.4\tiny{$\pm$0.3}       &79.0\tiny{$\pm$0.2}           &79.0\tiny{$\pm$0.2}    &79.5\tiny{$\pm$0.2}           &79.3\tiny{$\pm$0.3}       \\           
 VGG&CIFAR-10   &93.6\tiny{$\pm$0.2}       &93.5\tiny{$\pm$0.1}           &93.4\tiny{$\pm$0.1}&93.5\tiny{$\pm$0.1}           &93.5\tiny{$\pm$0.1}       \\   
 ResNet&CIFAR-10   &94.0\tiny{$\pm$0.2}       &93.8\tiny{$\pm$0.1}           &93.9\tiny{$\pm$0.1} &93.5\tiny{$\pm$0.3}           &93.5\tiny{$\pm$0.1}       \\\hline
 \toprule
 \multicolumn{7}{p{16cm}}{\textbf{Models.} FC4: Fully connected network with $4$ hidden layers, and $128$ units per layer. CONV4: Convolutional network with $4$ hidden layers, and $128$ filters per layer, followed by global-average-pooling, and output layer. VGG is VGG-16 and ResNet is ResNet-110 from \cite{gahaalt}. \textbf{Shuffling.} The last two columns contains the results after shuffling. Please look at \Cref{fig:c4gap} to see how to shuffle the layers in FC4 (DGN) and CONV4 (DGN). Please look at \Cref{sec:shuffle} to see how to shuffle VGG and ResNet.}\\
 \bottomrule
\end{tabular}
\caption{Entries in the first $3$ columns and the $3^{rd}$ row for VGG are averaged over $5$ independent runs. The entries in the last row for ResNet are averaged over $3$ independent runs. The entries in the first two rows and last two columns are averaged over the $23$ models obtained via permuting the layers (here each model is run once). The results in this table are in the setting wherein the gating network of the DGNs are pre-trained. However, the same results also hold even when the gating network is trained from scratch simultaneously alongside the weight network, and this is shown in \Cref{tb:dgntable-dlg}. In short, pre-training the gating network is not critical for the surprising results in the constant input and layer shuffling scenarios.}
\label{tb:dgntable}
\end{table*}

\section{Weight Network is Unnecessary for Interpretability}\label{sec:weightnetwork}
The theoretical understanding of the Neural Path Kernel in the previous section leads us to the following fundamental claim about Deep Gated Networks namely \emph{the weight network is unnecessary for interpretability}. We perform two sets of experiments to back this claim. These experimental conclusions would apriori seem counter intuitive and challenge the conventional understanding of what a network is learning. However, viewed with the insight gained from the theoretical results based on the dual view, these conclusions make sense and will lead us to propose a more interpretable architecture in the next section. 
\subsection{Constant Input Does Not Hurt Performance}
Our first experiment attacks the input to a standard DGN. A standard DGN uses the same input to both the gating network and weight network. In our modified set up, we only supply the input to the gating network and a constant all ones vector as input to the weight network. We refer to the resulting network architecture as DGN-ALLONES. From a conventional wisdom of `layer by layer' learning of sophisticated features, one would expect a significant drop in the performance when the input is made constant. However, we show in \Cref{tb:dgntable}, this is not the case. 

%We also consider the corresponding DGN counterparts. For details of the model architecture

%\input{alltables}
\begin{figure}[!ht]
\centering
\resizebox{0.5\columnwidth}{!}{
\begin{tikzpicture}
\node []  (dnn-text)at (5.375,2+2) {CONV4 (DNN)};

\node []  (dnn-output) at (9.25,1+2) {$\hat{y}(x)$};
\node []  (dnn-fc2) at (8.0,1+2) {\tiny{$FC$}};
\draw [-stealth,thick]   (dnn-fc2.east) -- (dnn-output.west);

\node [rotate=-90]  (dnn-gap) at (7.25,1+2) {\tiny{GAP}};
\draw [-stealth,thick]   (dnn-gap.north) -- (dnn-fc2.west);

\node [rotate=-90] (dnn-relu-4) at (6.5,1+2){\tiny{ReLU}};
\node [] (dnn-c4) at (5.75,1+2){{$C_4$}};
\draw [-stealth,thick]   (dnn-c4.east) -- (dnn-relu-4.south);
\draw [-stealth,thick]   (dnn-relu-4.north) -- (dnn-gap.south);

\node [rotate=-90] (dnn-relu-3) at (5,1+2){\tiny{ReLU}};
\node [] (dnn-c3) at (4.25,1+2){{$C_3$}};
\draw [-stealth,thick]   (dnn-c3.east) -- (dnn-relu-3.south);
\draw [-stealth,thick]   (dnn-relu-3.north) -- (dnn-c4.west);

\node [rotate=-90] (dnn-relu-2) at (3.5,1+2){\tiny{ReLU}};
\node [] (dnn-c2) at (2.75,1+2){{$C_2$}};
\draw [-stealth,thick]   (dnn-c2.east) -- (dnn-relu-2.south);
\draw [-stealth,thick]   (dnn-relu-2.north) -- (dnn-c3.west);

\node [rotate=-90] (dnn-relu-1) at (2,1+2){\tiny{ReLU}};
\node [] (dnn-c1) at (1.25,1+2){{$C_1$}};
\draw [-stealth,thick]   (dnn-c1.east) -- (dnn-relu-1.south);
\draw [-stealth,thick]   (dnn-relu-1.north) -- (dnn-c2.west);

\node [] (dnn-input) at (0.5,1+2){$x$};
\draw [-stealth,thick]   (dnn-input.east) -- (dnn-c1.west);

%%%%%%%%%%%%%%%%%%%%%%%%%%%%%%%%%%%%%%%%%%%%%%%%%%%%%%%%%%%%%%%%%

\node []  (fntext)at (5.375,2.25) {CONV4 (DGN) WITH SHUFFLING};

%\node []  (output) at (7.5,1.5) {$\hat{y}(x)$};

\node [] (dgn-f-c4) at (6.25,1.5){\tiny{$C^{\text{g}}_4$}};

\node [rotate=-90] (dgn-relu-3) at (5.5,1.5){\tiny{ReLU}};
\node [] (dgn-f-c3) at (4.75,1.5){\tiny{$C^{\text{g}}_3$}};
\draw [-stealth,thick]   (dgn-f-c3.east) -- (dgn-relu-3.south);
\draw [-stealth,thick]   (dgn-relu-3.north) -- (dgn-f-c4.west);

\node [rotate=-90] (dgn-relu-2) at (4.0,1.5){\tiny{ReLU}};
\node [] (dgn-f-c2) at (3.25,1.5){\tiny{$C^{\text{g}}_2$}};
\draw [-stealth,thick]   (dgn-f-c2.east) -- (dgn-relu-2.south);
\draw [-stealth,thick]   (dgn-relu-2.north) -- (dgn-f-c3.west);

\node [rotate=-90] (dgn-relu-1) at (2.5,1.5){\tiny{ReLU}};
\node [] (dgn-f-c1) at (1.75,1.5){\tiny{$C^{\text{g}}_1$}};
\draw [-stealth,thick]   (dgn-f-c1.east) -- (dgn-relu-1.south);
\draw [-stealth,thick]   (dgn-relu-1.north) -- (dgn-f-c2.west);

\node [] (dgn-f-input) at (0.75,1.5){$x^{\text{g}}$};
\draw [-stealth,thick]   (dgn-f-input.east) -- (dgn-f-c1.west);

\node []  (dgn-output) at (9.375,-2.5) {$\hat{y}(x)$};
\node [] (dgn-smax) at (8.375,-2.5){\tiny{FC}};
\draw [-stealth,thick]   (dgn-smax.east)--(dgn-output.west);

\node [rotate=-90] (dgn-gap) at (7.75,-2.5){\tiny{GAP}};
\draw [-stealth,thick]   (dgn-gap.north)--(dgn-smax.west);

\node [rotate=-90] (dgn-galu-4) at (7,-2.5){\tiny{GaLU}};
\draw [-stealth,thick]   (dgn-galu-4.north) -- (dgn-gap.south);

\node [] (dgn-v-c4) at (6.25,-2.5){\tiny{$C^{\text{w}}_4$}};
\draw [-stealth,thick]   (dgn-v-c4.east) -- (dgn-galu-4.south);

\node [rotate=-90] (dgn-galu-3) at (5.5,-2.5){\tiny{GaLU}};
\node [] (dgn-v-c3) at (4.75,-2.5){\tiny{$C^{\text{w}}_3$}};
\draw [-stealth,thick]   (dgn-v-c3.east) -- (dgn-galu-3.south);
\draw [-stealth,thick]   (dgn-galu-3.north) -- (dgn-v-c4.west);

\node [rotate=-90] (dgn-galu-2) at (4,-2.5){\tiny{GaLU}};
\node [] (dgn-v-c2) at (3.25,-2.5){\tiny{$C^{\text{w}}_2$}};
\draw [-stealth,thick]   (dgn-v-c2.east) -- (dgn-galu-2.south);
\draw [-stealth,thick]   (dgn-galu-2.north) -- (dgn-v-c3.west);

\node [rotate=-90] (dgn-galu-1) at (2.5,-2.5){\tiny{GaLU}};
\node [] (dgn-v-c1) at (1.75,-2.5){\tiny{$C^{\text{w}}_1$}};

\draw [-stealth,thick]   (dgn-v-c1.east) -- (dgn-galu-1.south);
\draw [-stealth,thick]   (dgn-galu-1.north) -- (dgn-v-c2.west);

\node [] (dgn-input) at (0.75,-2.5){$x^{\text{w}}$};
\draw [-stealth,thick]   (dgn-input.east) -- (dgn-v-c1.west);

\node[] (dgn-gating-1-up) at (2.5,0.5){\tiny{$G_{1}$}};
\draw [-stealth,thick]   (dgn-f-c1.east) to[out=-90,in=90] (dgn-gating-1-up.north);

\node[] (dgn-gating-2-up) at (4,0.5){\tiny{$G_{2}$}};
\draw [-stealth,thick]   (dgn-f-c2.east) to[out=-90,in=90] (dgn-gating-2-up.north);

\node[] (dgn-gating-3-up) at (5.5,0.5){\tiny{$G_{3}$}};
\draw [-stealth,thick]   (dgn-f-c3.east) to[out=-90,in=90] (dgn-gating-3-up.north);

\node[] (dgn-gating-4-up) at (7,0.5){\tiny{$G_{4}$}};
\draw [-stealth,thick]   (dgn-f-c4.east) to[out=-90,in=90] (dgn-gating-4-up.north);

\node[] (dgn-gating-1) at (2.5,-1.5){\tiny{$G_{i_1}$}};
\draw [-stealth,thick]   (dgn-gating-1.south) -- (dgn-galu-1.west);

\node[] (dgn-gating-2) at (4,-1.5){\tiny{$G_{i_2}$}};
\draw [-stealth,thick]   (dgn-gating-2.south) -- (dgn-galu-2.west);

\node[] (dgn-gating-3) at (5.5,-1.5){\tiny{$G_{i_3}$}};
\draw [-stealth,thick]   (dgn-gating-3.south) -- (dgn-galu-3.west);

\node[] (dgn-gating-4) at (7,-1.5){\tiny{$G_{i_4}$}};
\draw [-stealth,thick]   (dgn-gating-4.south) -- (dgn-galu-4.west);

\node[] (permutation) at (4.75,-0.5){\small{Layer Permutation}};

\draw [-]   (dgn-gating-1-up.south) -- (permutation.north);
\draw [-]   (dgn-gating-4-up.south) -- (permutation.north);
\draw [-]   (dgn-gating-2-up.south) -- (permutation.north);
\draw [-]   (dgn-gating-3-up.south) -- (permutation.north);

\draw [-]  (permutation.south) --  (dgn-gating-1.north)  ;
\draw [-]  (permutation.south) --  (dgn-gating-2.north)  ;
\draw [-]  (permutation.south) --  (dgn-gating-3.north)  ;
\draw [-]  (permutation.south) --  (dgn-gating-4.north)  ;

\end{tikzpicture}
}
\caption{Top: CONV4 (DNN). Bottom: CONV4 (DGN) along with mechanism to shuffle the gating masks. Here $x^{\text{g}}$ and $x^{\text{w}}$ denote the input to the gating and weight network respectively. Here the gates $G_1, G_2, G_3, G_4$ are generated by the gating network and are permuted as $G_{i_1},G_{i_2},G_{i_3},G_{i_4}$ before applying to the weight network. Shuffling of FC4 (DGN) is similar to CONV4 (DGN). There are $24$ possible permutations of the gating masks. For FC4 and CONV4, the results of the models with identity permutation of layers are in column \texttt{WITHOUT SHUFFLING} in \Cref{tb:dgntable} and the results of the $23$ other models are in the column \texttt{WITH SHUFFLING} in \Cref{tb:dgntable}.}
\label{fig:c4gap}
\end{figure}
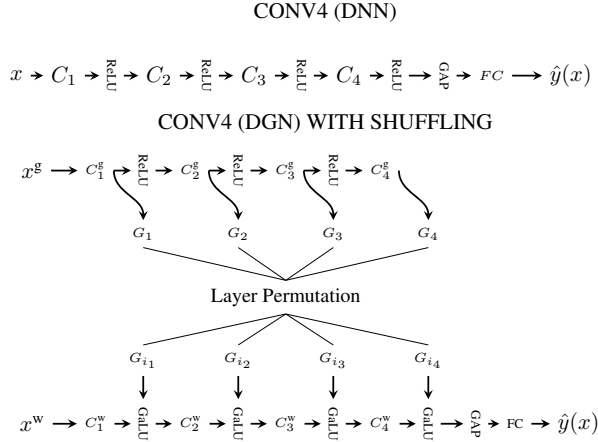

\textbf{Theoretical Justification:}
We will now explain how the results in \Cref{tb:dgntable} follows from results in the previous section. In the fully connected case, the expression on right hand side of the NPK expression in \Cref{th:fc} becomes $\Pi_{l=1}^{d-1}{\ip{G_l(x),G_l(x')}}$, i.e., the kernel still has information about the input encoded via the gates. In the case of convolutional networks, the expression in \Cref{th:conv} becomes $\cdot \sum_{r=0}^{\din-1}\sum_{i=1}^{\din} \textbf{overlap}(i,x,rot(x',r))$. The key novel insight is that the rotational invariance is not lost and $\textbf{overlap}$ matrix measures the correlation of the paths which in turn depends on the correlation of the gates. In the case of ResNets, as the $\text{NPK}^{\texttt{RES}}$ of the residual networks is an ensemble, the property of the block level kernel (i.e., of $\text{NPK}^{\texttt{FC}}$/$\text{NPK}^{\texttt{CONV}}$) translates to $\text{NPK}^{\texttt{RES}}$.

\subsection{Layer Shuffling Does not Hurt Performance}
Our second experiment attacks the hidden layers of the weight network of standard DGN. In addition to the all ones input, we now also randomly shuffle the hidden layers i.e., the gating masks are applied post layer shuffling (See \Cref{fig:c4gap}). Again, if the conventional wisdom about layer-by-layer learning were true, then the performance after layer shuffling must take a big hit. We show in \Cref{tb:dgntable} that this is not the case.

\textbf{Theoretical Justification:}
Recall that in the NPK expression for the fully connected case in \Cref{th:fc}, $\Pi_{l=1}^{d-1}\ip{G_l(x),G_l(x')}$ is permutation invariant, and hence permuting the layers has no effect, in that, it leaves the NPK expression unchanged. Similarly, permuting the layers does not destroy the rotational invariance in \Cref{th:conv}. This is because, due to circular convolutions all the internal variables of the network rotate as the input rotates. Permuting the layers only affects the ordering of the layers, and does not affect the fact that the gates rotate if the input rotates. As explained previously, ResNet being an ensemble of block level kernels, inherits the permutation invariance property from the blocks (be it fully connected or convolutional).

\textbf{Remark:} In the case of ResNets, even removing layers does not hurt performance. The ensemble behaviour of ResNet and  presence of $2^b$ architectures was observed by \cite{veit2016residual}, however without any concrete theoretical formalism. \cite{veit2016residual} showed empirically that removing single layers from ResNets at test time does not noticeably affect their performance, and yet removing a layer from architecture such as VGG leads to a dramatic loss in performance. \Cref{th:res} can be seen to provide a theoretical justification for this empirical result. The ResNet inherits the invariances of the block level kernel. In addition, the ensemble structure allows to even remove layers. In other words, due to the ensemble structure a ResNet is robust to failures, in particular, the insight is that even if one or many of the kernels in the ensemble are corrupt, the good ones can compensate.

\textbf{Remark:} A qualitative justification of the above results can be found in \Cref{fig:horse} (see caption for details).

\begin{figure*}
\resizebox{1.0\columnwidth}{!}{
\begin{tabular}{rcl|c|c|c|c|}\cline{4-7}
 & & &\multicolumn{2}{c|}{VGG (CIFAR-10: 93.6{\tiny{$\pm$0.2}}, CIFAR-100:73.4{\tiny{$\pm$0.3}} )}&\multicolumn{2}{c|}{ResNet(CIFAR-10: 94.0{\tiny{$\pm$0.2}}, CIFAR-100:72.7{\tiny{$\pm$0.2}} )}\\\cline{4-7}
&&&CIFAR-10&CIFAR-100&CIFAR-10&CIFAR-100\\\hline
\multicolumn{1}{|r}{SHALLOW-LIN} &$\stackrel{\text{dual lift}}{\longrightarrow}$&DLGN-SHALLOW&30.6{\tiny{$\pm$0.5}} $\rightarrow$ 84.9{\tiny{$\pm$0.2}} &10.5{\tiny{$\pm$0.2}} $\rightarrow$ 56.3\tiny{$\pm$0.2}&30.6{\tiny{$\pm$0.5}} $\rightarrow$ 80.1\tiny{$\pm$0.2}&10.5{\tiny{$\pm$0.2}} $\rightarrow$49.8\tiny{$\pm$0.9}\\
\multicolumn{1}{|r}{DEEP-LIN} &$\stackrel{\text{dual lift}}{\longrightarrow}$&DLGN&37.8{\tiny{$\pm$0.5}} $\rightarrow$  87.0\tiny{$\pm$0.1}        &15.4{\tiny{$\pm$0.5}} $\rightarrow$ 61.5\tiny{$\pm$0.2} &38.9{\tiny{$\pm$0.1}}   $\rightarrow$    87.9\tiny{$\pm$0.2}    &16.9{\tiny{$\pm$0.1}}$\rightarrow$62.5\tiny{$\pm$0.2}\\
\multicolumn{1}{|r}{SHALLOW-ReLU} &$\stackrel{\text{dual lift}}{\longrightarrow}$&DLGN-SHALLOW& 62.2{\tiny{$\pm$0.2}}$\rightarrow$  84.9{\tiny{$\pm$0.2}}      &36.3{\tiny{$\pm$0.2}}$\rightarrow$ 56.3\tiny{$\pm$0.2} &62.2{\tiny{$\pm$0.2}} $\rightarrow$ 80.1\tiny{$\pm$0.2}       &36.3{\tiny{$\pm$0.2}} $\rightarrow$49.8\tiny{$\pm$0.9} \\
\multicolumn{1}{|r}{DEEP-LIN-MAX} &$\stackrel{\text{dual lift}}{\longrightarrow}$&DLGN-MAX&88.3{\tiny{$\pm$0.1}}$\rightarrow$  92.1\tiny{$\pm$0.1}         &61.8{\tiny{$\pm$0.5}}$\rightarrow$ 69.1\tiny{$\pm$0.1} &84.6{\tiny{$\pm$0.2}}   $\rightarrow$  89.5\tiny{$\pm$0.3}      &57.0{\tiny{$\pm$0.1}}$\rightarrow$66.9\tiny{$\pm$0.1}\\\hline
\end{tabular}
}
\begin{minipage}{0.25\columnwidth}
\resizebox{1.05\columnwidth}{!}{
\begin{tikzpicture}

%%%%%%%%%%%%%%%%%%%%%%%%%%%%%%%%%%
\node [color=red]  (dlgn-fntext)at (5+5,1+0.25) {\tiny{\bf{Deep Linearly Gated Network}}};
%Feature Network
\node [draw,
	minimum width=2cm,
	minimum height=0.625cm,
]  (dlgn-fnbox)at (5+5,0.375+0.25) {};
\node []  (dlgn-fntext)at (5+5,0.5+0.25) {\tiny{Gating Network}};

\node []  (dlgn-fntext)at (5+5,0.25+0.25) {\tiny{\bf\color{red}{Deep Linear Network}}};

%Feature Network Input
\node (dlgn-fin) [left of=dlgn-fnbox,node distance=1.25cm, coordinate] {};
\node[left=-1pt] at (dlgn-fin.west){\tiny{Input}};
\draw[-stealth] (dlgn-fin.center) -- (dlgn-fnbox.west);

%Value Network

\node [draw,
	minimum width=2cm,
	minimum height=0.625cm,
]  (dlgn-vnbox)at (5+5,-0.875-1.5) {};
\node []  (dlgn-fntext)at (5+5,-0.75-1.5) {\tiny{Weight Network}};
\node []  (dlgn-vntext)at (5+5,-1-1.5) {\tiny{DNN with GaLUs}};

%Value Network Input
\node (dlgn-vin) [left of=dlgn-vnbox,node distance=1.25cm, coordinate] {};
\node[left=-1pt] at (dlgn-vin.west){\tiny{\color{red}{Input =$\mathbf{1}$}}};
\draw[-stealth] (dlgn-vin.center) -- (dlgn-vnbox.west);

%Feature Network Output
\node (dlgn-vout) [right of=dlgn-vnbox,node distance=1.25cm, coordinate] {};
\node[right=-1pt] at (dlgn-vout.west){\tiny{Output}};
\draw[-stealth]  (dlgn-vnbox.east)--(dlgn-vout.center);

\node []  (dlgn-preacr)at (5.75+5,0.125-0.25) {\tiny{Pre-activations}};

\node []  (dlgn-gating)at (5.75+5,-0.375-1) {\tiny{Gating Signal}};

\node []  (dlgn-gates)at (5.0+5,-0.125-0.75) {\color{red}\tiny{Gates}};

\node [draw,
	minimum width=1cm,
	minimum height=0.1cm,
]  (dlgn-gbox)at (5+5,-0.125-0.75) {};

\draw[-stealth]  (dlgn-gbox.south)--(dlgn-vnbox.north);
\draw[-stealth]  (dlgn-fnbox.south)--(dlgn-gbox.north);

\end{tikzpicture}
}
%\begin{tabular}{c}\\\\\\\\\hline
%\multicolumn{1}{|p{4cm}|}{In a DLGN, both gating and weight network are %trained simultaneously (for details please see %\Cref{sec:dgn-training}).}\\\hline
%\end{tabular}
\end{minipage}
\begin{minipage}{0.7\columnwidth}
\resizebox{1.0\columnwidth}{!}{
\includegraphics[scale=0.29]{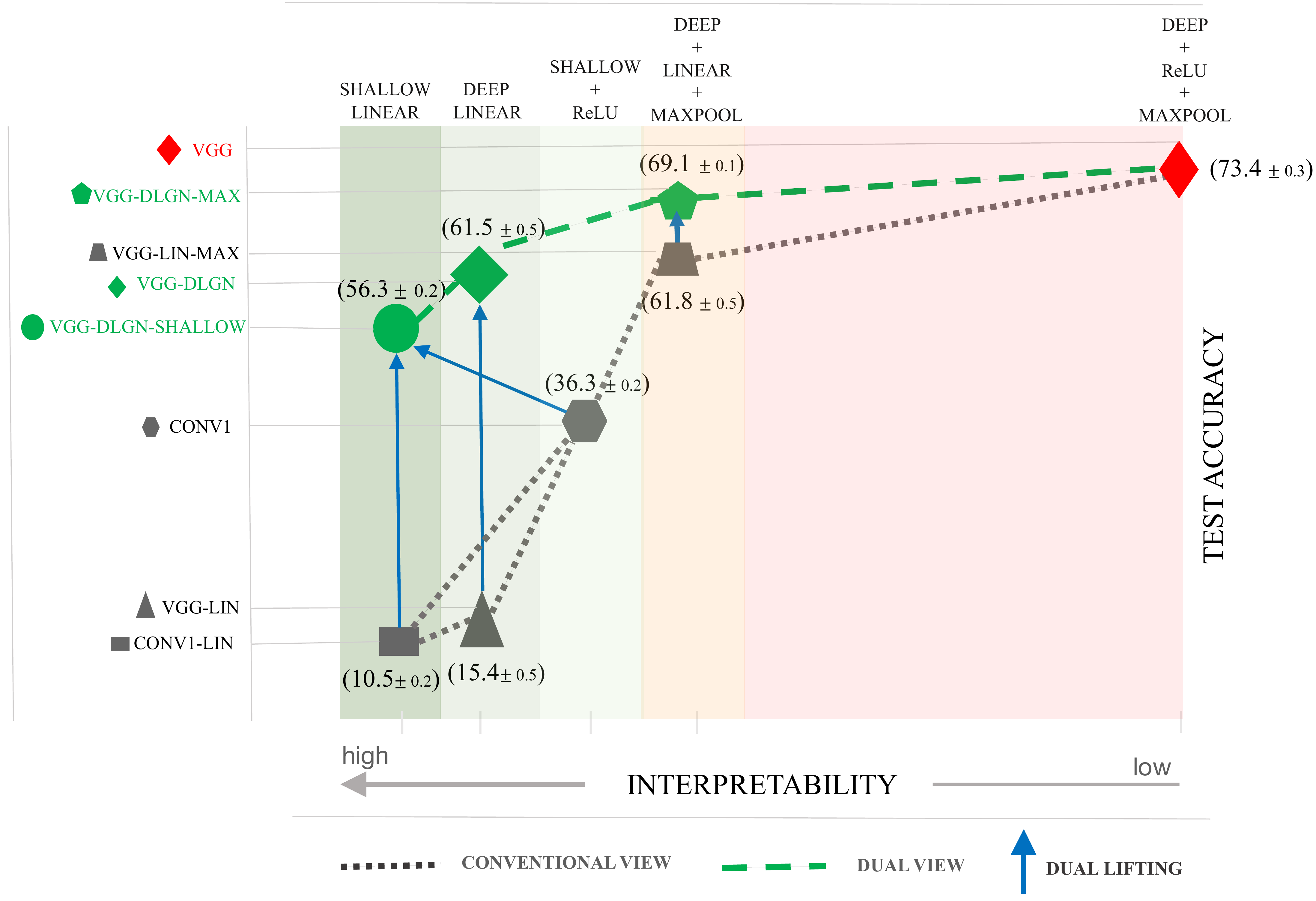}
}
\end{minipage}
\caption{\textbf{Left:} Shows the DLGN which is DGN whose gating network is linear. In the DLGN, both the gating and weight networks are trained simultaneously using the same protocol (step size, learning rate, optimiser) followed to train the corresponding DNNs. \textbf{Right:} Shows the interpretability-accuracy tradeoff plot for the various models considered. The gray path shows models that are obtained via the conventional view, and the green path shows models obtained via DLGN (proposed in this paper) by dual lifting (see \Cref{sec:duallift}). \textbf{Top:} Shows the tabular column of test accuracies of all the models. Here, the results for ResNet variants are averaged over 3 independent runs and the results for other models are averaged over 5 independent runs.}
\label{fig:road}
\end{figure*}

\subsection{Dual Lifting}\label{sec:duallift}
The above experiments lead to the important conclusion that the non-linear weight network of DGNs while crucial for ensuring good performance, must not be used to interpret what is being learnt in the layers. Then, a natural question is to understand the role of the weight network. This is easy to answer from the dual viewpoint, paying attention to the phenomena of what we call as \emph{dual lifting}, which we explain below. 

The gating network provides the pre-activations to the GaLUs in the weight network. By switching on/off the GaLUs, the paths in the weight network are turned on/off. This causes the \textbf{dual lifting} of the features in the form of layer outputs from the gating network to get \textbf{lifted} to a neural path feature vector, i.e., a vector in the dimension of the total number of paths. Now, the neural path value is a function of the weights in the weight network and finally the \emph{dual linearity} i.e., the inner product between the neural path feature and neural path value is computed. In short, dual lifting causes the information in the layers of the gating network to be thrown into the dimension of paths.

\subsection{Implications}
As a natural consequence of the fact that the weight network is not needed for interpretation, any attempt at interpreting DGNs must focus on what is being learnt in the gating network. However, DGNs have non-linearity (ReLUs) in the gating network which make interpretations harder. A natural question is then \emph{Is the ReLU in the Gating Network crucial for good performance?}. Surprisingly, the answer seems to be a no. We discuss this in the next section in detail. 

\section{Rethinking Interpretability via Deep Linearly Gated Networks}
The experimental results of the previous section rule out the necessity for looking at the weight network of the DGN for interpretability. Thus we turn our attention to the gating network. To make DGNs interpretable, we propose the Deep Linearly Gated Network (DLGN) (see left hand side of \Cref{fig:road}) wherein we completely eliminate the non-linearity of the gating network by removing the ReLU activations. Before we explain the results that we obtain for DLGNs, we take a step back and discuss interpretability in general. This will help us contextualise our results better.

The simplest \emph{interpretable} model is the shallow linear model. In this model, while retaining linearity, one can increase it's complexity by introducing deep layers. Instead of adding deep layers, if one adds ReLU to a shallow linear model, one gets the well studied ReLU networks with single hidden layer \cite{chizat2018global,zhang2019learning, li2018learning,zhong2017recovery, li2017convergence, ge2017learning,du2018cnn}. However, the non-linearity in the ReLU makes it less interpretable compared to a deep linear model. As a next step, one might want to add the standard deep learning \emph{tricks} such as striding, average/max-pooling and batch normalisation to a deep linear model. Out of these tricks, only max-pooling is non-linear. Yet, interpretability wise it may be considered a benign form of non-linearity in comparison to ReLU, the reason being that, max-pooling acts directly on the input as opposed to ReLU which acts on the input transformed by a weight vector. As max-pooling is essentially a sub-sampling operation a deep linear model with max-pooling could potentially be interpreted by understanding the linear part and max-pooling separately. Once non-linearity is introduced in the form of say ReLU in deep networks, one typically loses interpretability. Thus one can arrange the increasing levels of interpretable models as follows:

\begin{align}\label{eq:order}
\begin{split}
&\texttt{Shallow Linear} > \texttt{Deep Linear} > \\
&\texttt{Shallow + ReLU} > \texttt{Deep Linear + MaxPool}  >\\
&\texttt{Deep + ReLU} > \texttt{Deep + MaxPool + ReLU}
\end{split}
\end{align}
The conventional viewpoint is that as we gain interpretability, we lose accuracy in proportional measure. In practice then one resorts to choosing accuracy over interpretability and then attempts to interpret/explain the model in a post-hoc fashion \cite{lime,gradcam,gradcampp}. However, several recent works have pointed out the need for developing built-in interpretable models i.e., making deep models more interpretable and avoiding taking a post-hoc view of the same \cite{rudinstop,rudininterpretable}. We argue that this is possible by showcasing such a model. In fact, one can still have \emph{essentially linear} interpretable models which perform almost as well as DNNs/DGNs thanks to the phenomenon of \emph{Dual Lifting} in weight network. The main idea is that we start with a standard DGN and make two key changes: (i) we make the gating network linear and (ii) we give all ones (non-useful) input to the weight network. We refer to this model a Deep Linearly Gated Network. The network still has GaLU activations as in DGNs but only in the weight network which we already know from  \Cref{sec:weightnetwork} does not play a role in interpretability. Thus, non-linearity is present to perform a \emph{lifting} from the primal space to the dual space. By clearly separating the linear part from the non-linear part and noting that the non-linear part does not play a role in interpretability, we gain the advantages of both worlds. 

We describe our results in \Cref{fig:road} (the table in the top and the plot in the bottom right). The figure lays out the landscape of interpretable models starting from the shallow linear model all the way upto the standard Deep + Maxpool + ReLU models. We first describe the models considered in detail below. 

\textbf{Models Considered:} Our aim is to create one conventional model for each of the cases in the inequality in \Cref{eq:order} and the corresponding DLGN counterpart. CONV1 has 1 convolutional layer with 512 filters, followed by ReLU, global-average-pooling and output layer (this is the shallow + ReLU model). CONV1-LIN is same as CONV1 with the ReLU activations replaced by identity activation (this is the shallow linear model).  VGG-LIN is same as VGG with average pooling instead of max-pooling (to ensure linearity) and ReLUs replaced by identity activation. VGG-LIN-MAX is VGG (i.e., with max-pooling) and ReLUs replaced by identity activation. VGG-DLGN is the DLGN counterpart of VGG (the architecture is in \Cref{fig:vgg_dlgn} in the appendix). VGG-DLGN-SHALLOW is obtained by modifying the gating network of VGG-DLGN as follows: the pre-activations to the gates are generated by $d$ shallow linear convolutional networks (the architecture is in \Cref{fig:vgg_dlgn_shallow} in the appendix).VGG-DLGN-MAX is same as VGG-DLGN, however with max-pooling instead of average pooling.

Conventional thinking to understand interpretability-accuracy tradeoff would suggest increasing model complexity one step at a time starting right from shallow linear networks i.e., moving along one of the gray paths in \Cref{fig:road}. The proposed model, DLGN and it's corresponding variants for each of the cases, by exploiting \emph{Dual Lifting}, outperforms the conventional models at every stage. This corresponds to the green path in \Cref{fig:road}. By taking the green path, one retains as much interpretability as their counterpart models in the conventional way of thinking while still gaining accuracy. This is the primary advantage of DLGNs.

\textbf{Lifting:}
It is interesting to note that there is a significant gain due to lifting from the Shallow Linear Model to the  DLGN-Shallow model. The gain decreases as one adds deep layers followed by MaxPool. This is expected as the effect of the non-linear, non-interpretable weights network is most pronounced when the original model that it is compared against is linear (and shallow). As one introduces some form of non-linearity (MaxPool or ReLU), the lifting helps lesser. %We delve deeper and explain why lifting works in the next Section (Section \ref{}).

\textbf{Deep Linear + MaxPool:} We observe that adding max-pool to deep linear models results in significant boost in performance (VGG/ResNet-LIN-MAX performs better than VGG/ResNet-LIN). It is worth noting that dual lifting of VGG-LIN-MAX to VGG-DLGN-MAX further boosts the performance achieving accuracies of $\mathbf{69.1\%}$ and $\mathbf{92.1\%}$ on CIFAR-10 and CIFAR-100 respectively, which are within $\mathbf{4.3\%}$ and $\mathbf{1.5\%}$ of the performance of VGG on the respective datasets. As discussed before, VGG-DLGN-MAX is worthy of further interpretation by seperately understanding the linear part and max-pooling. Also, it would be an interesting future study to understand the reason for the boost in performance when max-pooling is added to a deep linear model.

\textbf{Deep Linear and Shallow Linear:} While the representation power of the deep and shallow linear networks is the same, the difference between them is that of optimisation. We observe that both these models perform poorly in both CIFAR-10 as well as CIFAR-100. Since our main focus has been in interpretable models that also achieve reasonable test accuracy, we did not probe further into these two models. Nevertheless, deep linear networks have been studied well in the literature, especially in connection to the optimisation of deep networks \cite{shamir,dudln,ganguli,ji2018gradient}.

\textbf{SHALLOW-ReLU vs DLGN-SHALLOW:} Note that the gating network of the DLGN-SHALLOW comprises of several shallow linear networks in parallel, each which in turn trigger the gates of a layer in the corresponding weight network. Thus if one dual lifts a SHALLOW-ReLU model we obtain a DLGN-SHALLOW model (as seen in \Cref{fig:road}). We observe that VGG-DLGN-SHALLOW performs better than CONV1. Note that VGG-DLGN-SHALLOW is better than CONV1 both in terms of interpretability and accuracy.

\textbf{Dual Lifting and Standard DNNs:} Given the performance gain due to dual lifting, a natural question to ask is whether dual lifting is a phenomena solely restricted to DGN/DLGNs or can it also be used to improve the performance of standard DNNs. This is answered by looking at standard DNNs with ReLUs to be models that have their gating and weight networks to be one and the same, and since dual lifting happens in the weight network, it follows that \emph{dual lifting is implicit in standard DNNs with ReLUs}. Seen this way, one can attribute the success of standard DNNs with ReLUs also to dual lifting.

\textbf{Remark:} The trend for the ResNet variants also follows in a similar manner as VGG variants (see table on top of \Cref{fig:road}). We also observe that the performance gap between the various models is less on CIFAR-10 and more pronounced in CIFAR-100, which is expected since CIFAR-100 (with $100$ classes) is a harder dataset than CIFAR-10 (with $10$ classes).

%\subsection{Rethinking Interpretability}
%DLGN are in some sense  are 'minimally' non-linear since the gating network is forced to be linear. The experimental success of DLGNs lead us to believe that DLGNs provides a new way of thinking about interpretability of deep neural networks. Any deep learning practioner who wishes to \emph{interpret} what the network is learning using simple (linear) models must thus look at what is being learnt in the linear gating network. We believe that domain specific conclusions about what is being learnt (eg: image processing filters, etc) can be arrived at as this model is used for specific tasks. We leave this domain specific interpretations to future work.

%\textbf{A Note on Further Experiments}:
%In \cite{npk}, DGNs were tested on several experimental settings including {\color{red} list the settings}. We have done experiments on all these settings and the main message is the same - The primal-dual lifting helps  Due to lack of space, we report the primary settings in the main paper and refer the reader to the supplementary material for details of the remaining settings. 

%\section{Lifting Gives Rise To Composite Kernel Learning}
%\section{DLGN for Continual Learning}
%DLGNs have a conceptual similarity to the Gated Linear Networks (GLN) of \cite{gln}.  GLNs have external gating similar to DGN/DLGN but the gating is not learnt. It is known that GLNs perform well on continual learning tasks. As a concrete application of DLGNs, we wish to see how DLGNs can be adapted to the task of learning tasks as well. 
\section{Conclusion}
In this paper, we rethink the conventional interpretations of DNNs by taking a dual view. By exploiting novel theoretical properties of Neural Path Kernel that we identify, we demonstrate via extensive experiments that  only the Gating network of a Deep Gated Network is necessary for interpretability. We propose Deep Linearly Gated Networks, an interpretable counterpart to DGNs and show how it makes use of the phenomenon of dual lifting to improve accuracy over conventional approaches while maintaining interpretability. 

We believe our results on the surprisingly good performance of DLGNs (as compared to DGNs) would motivate DNN researchers to take a closer look at domain specific understanding of DLGNs. Future work includes understanding DLGNs from the point of view of adversarial attacks.

%To elaborate, the feature network provides the gates $G_l(x)$, and the value network realises the product kernel in \Cref{th:fc} by laying out the GaLUs depth-wise, and connecting them to form a deep network. The depth-wise layout is important: for instance, if we were to concatenate the gating features as $\varphi(x)=(G_l(x),l=1,\ldots,d-1)\in\{0,1\}^{(d-1)w}$, it would have only resulted in the kernel $\ip{\varphi(x),\varphi(x')}=\sum_{l=1}^{d-1}{\ip{G_l(x),G_l(x')}}$, i.e., a \emph{sum  (not product)} of kernels. 

\bibliographystyle{unsrtnat}
\bibliography{refs} 
\appendix
\section{Shuffling in VGG and ResNet}\label{sec:shuffle}
\FloatBarrier
\begin{figure}[h]
    \centering
    \includegraphics[scale=0.25]{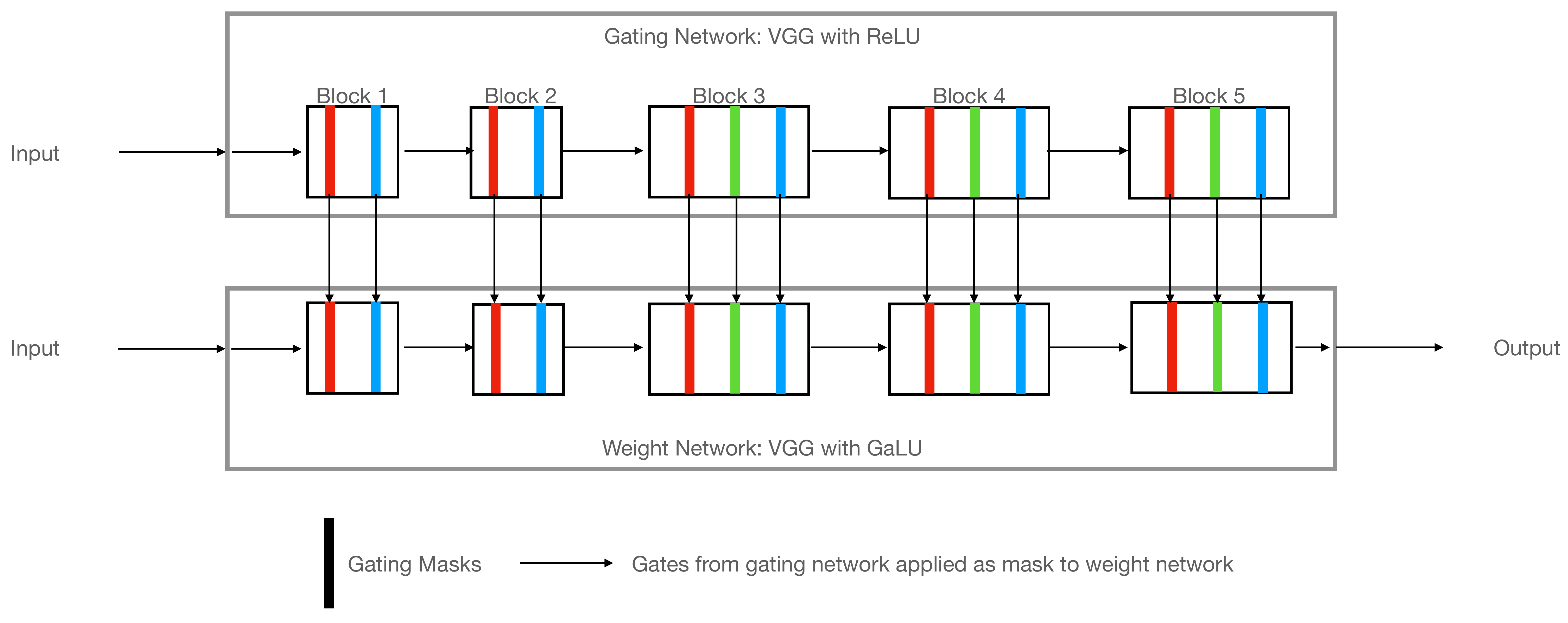}
    \caption{Shows VGG (DGN).}
    \label{fig:vgg_dgn}
\end{figure}

\begin{figure}[h]
    \centering
    \includegraphics[scale=0.25]{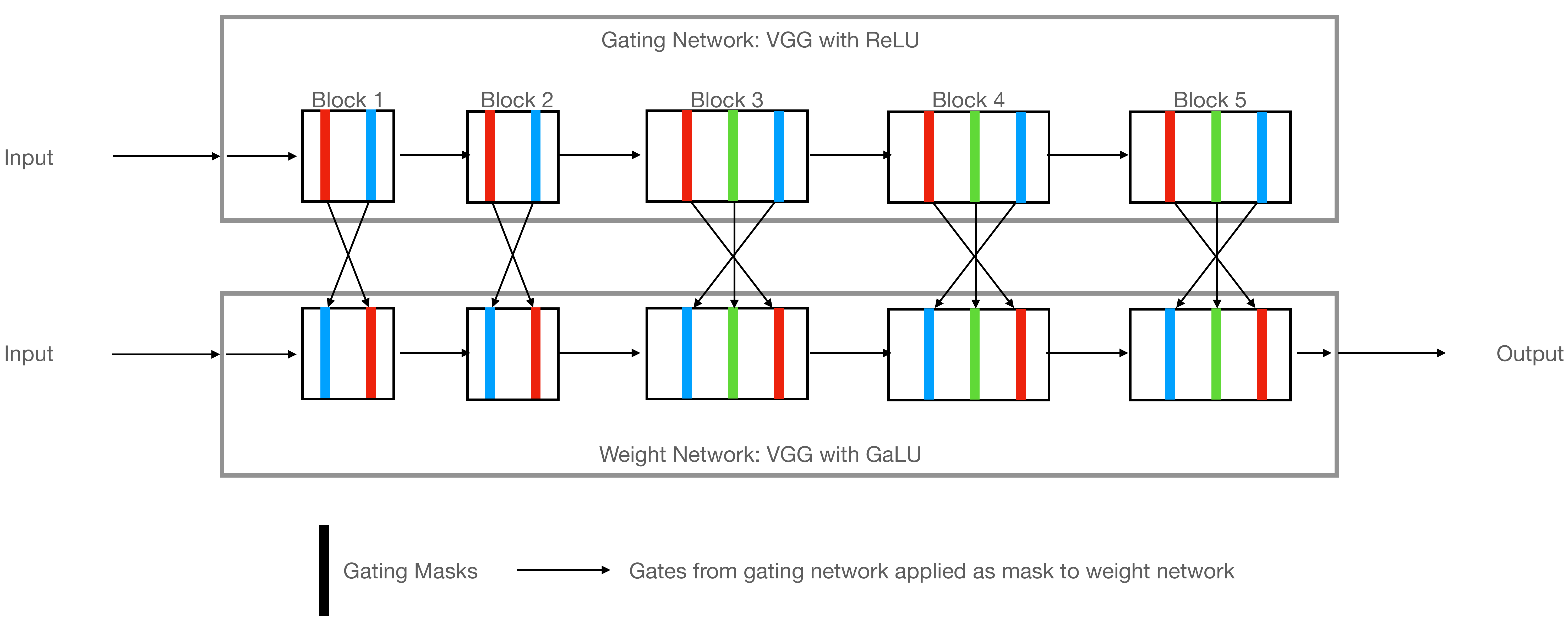}
    \caption{Shows shuffling of gating masks in VGG (DGN) in the last two columns, $3^{rd}$ row in \Cref{tb:dgntable}. The shuffling of ResNet (DGN) is similar. VGG contains $5$ blocks each with filter sizes (64,128,256,512,512) and number of layers within a block (2,2,3,3,3) respectively. Since the number of filters differ across the blocks we shuffle the gating of the layers within each block in the reverse order.ResNet contains $3$ blocks each with filter sizes (16,32,64) and number of layers within a block (36,36,36) respectively. The shuffling is done in a similar manner as VGG,i.e., the gating masks are reversed with each block.}
    \label{fig:vgg_shuffle}
\end{figure}
\newpage

\subsection{VGG-DLGN}
\FloatBarrier
\begin{figure}[h]
    \centering
    \includegraphics[scale=0.25]{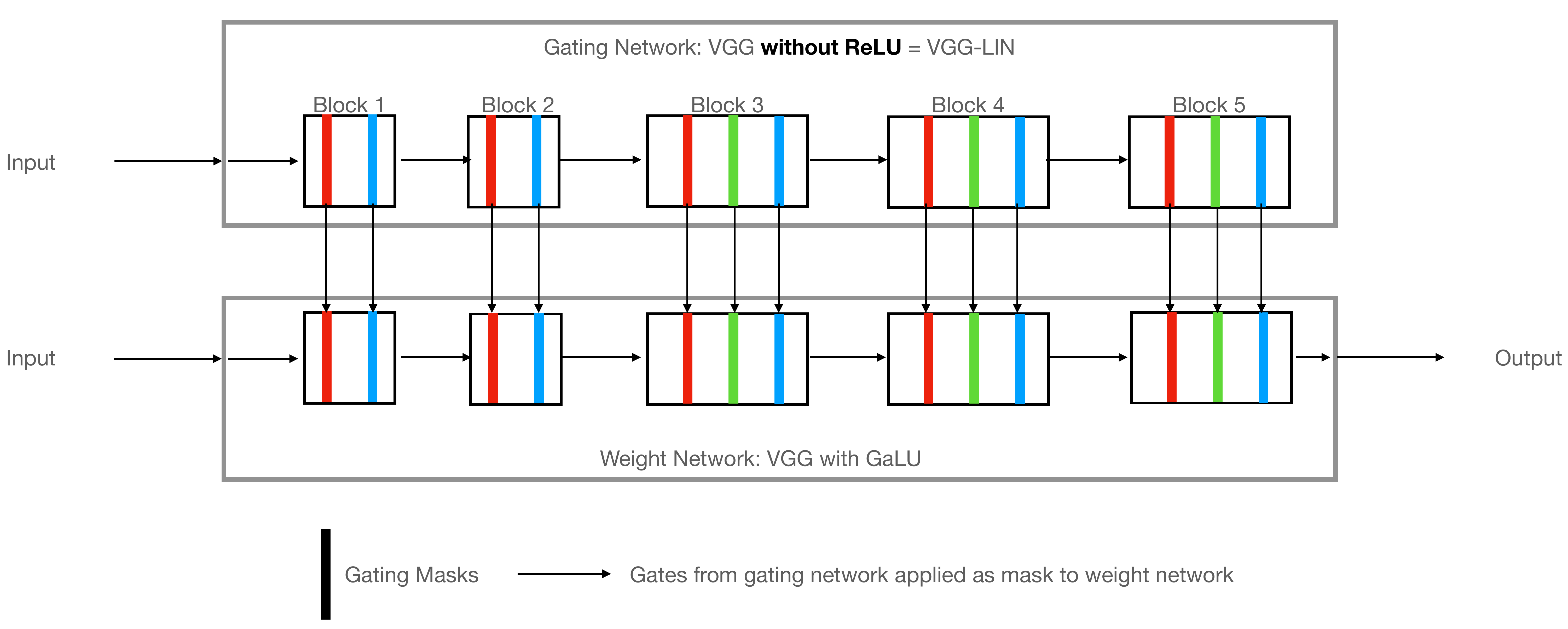}
    \caption{Shows VGG-DLGN}
    \label{fig:vgg_dlgn}
\end{figure}

\subsection{VGG-DLGN-SHALLOW}
\FloatBarrier
\begin{figure}[h]
    \centering
    \includegraphics[scale=0.25]{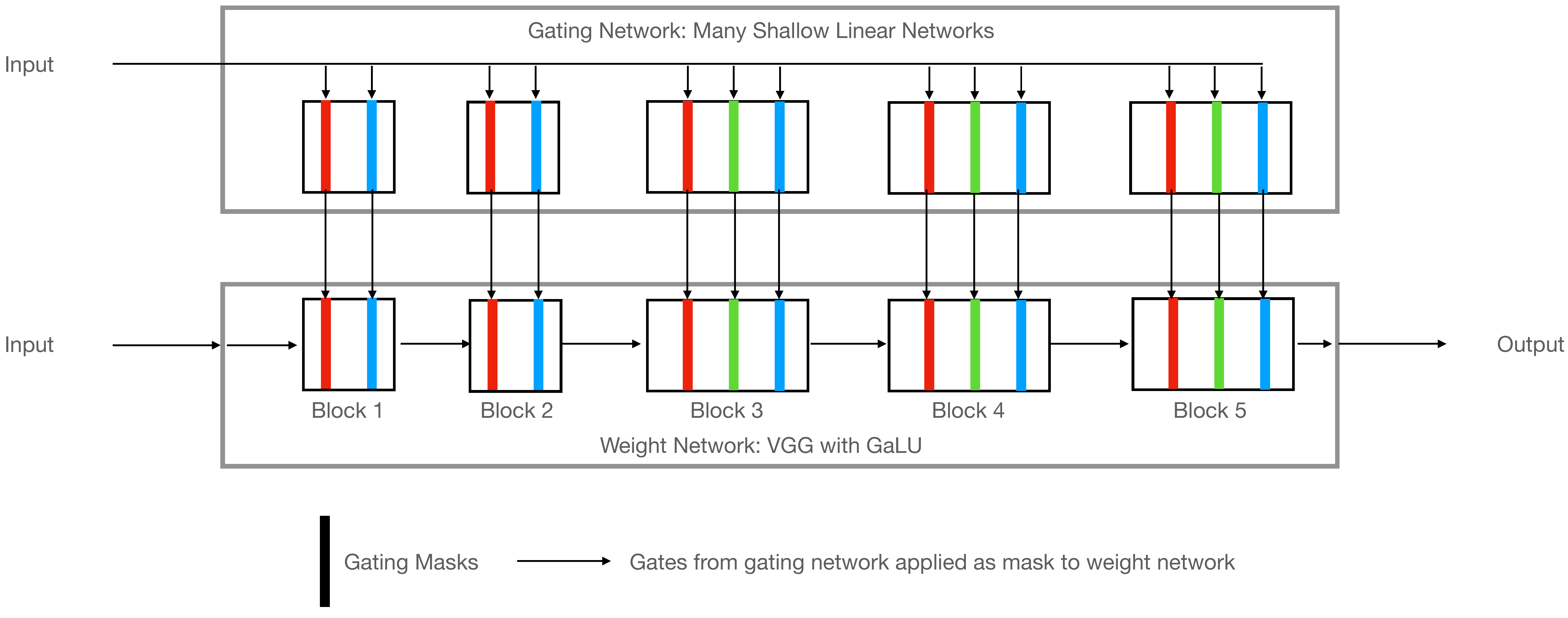}
    \caption{Shows VGG-DLGN-SHALLOW.}
    \label{fig:vgg_dlgn_shallow}
\end{figure}
\newpage
\section{DGN Results without pre-training of the gating network}
\FloatBarrier
\begin{table}[h]
\centering
\begin{tabular}{|c|c|c|c|c|c|c|}\cline{4-7}
\multicolumn{3}{c}{}&\multicolumn{2}{|c|}{\texttt{WITHOUT SHUFFLING}} &\multicolumn{2}{c|}{\texttt{WITH SHUFFLING}} \\\hline
Architecture & Dataset & DNN & DGN & DGN-\texttt{ALLONES} & DGN & DGN-\texttt{ALLONES}\\\hline
 FC4            &MNIST      &98.5\tiny{$\pm$0.1}       &98.2\tiny{$\pm$0.1}           &98.2\tiny{$\pm$0.1}       &98.1\tiny{$\pm$0.1}           &98.2\tiny{$\pm$0.1}    \\        
 CONV4          &CIFAR-10   &80.4\tiny{$\pm$0.3}       &77.2\tiny{$\pm$0.4}          &77.7\tiny{$\pm$0.3} &77.3\tiny{$\pm$0.5}           &77.8\tiny{$\pm$0.6}      \\   
 VGG         &CIFAR-10   &93.6\tiny{$\pm$0.2}       &93.0\tiny{$\pm$0.1}           &93.0\tiny{$\pm$0.1}    &92.9\tiny{$\pm$0.1}           &92.9\tiny{$\pm$0.3}       \\           
 ResNet     &CIFAR-10   &94.0       &93.3\tiny{$\pm$0.2}           &93.3\tiny{$\pm$0.1}      &92.9\tiny{$\pm$0.1}    &92.9\tiny{$\pm$0.2} \\\hline         
\end{tabular}
\caption{\Cref{tb:dgntable} showed the results when the gating network was pre-trained. This table shows the results when the gating network of the DGN is trained from scratch simultaneously alongside the weight network. The drop in $2-3\%$ of the test accuracy when the gating network of teh DGNs are trained from scratch as opposed to a DGN with pre-trained gates has been documented by \cite{npk} and we observe that the same holds here as well. However, the most critical claim of this paper that providing a constant input as well as shuffling the gating masks does not \textbf{further} degrade the performance continues to hold even in the case when the gating network is trained from scratch is what is shown in this table. In other words, evidence provided by \Cref{tb:dgntable} to support our critical claims is not dependent on the fact that the gating network was pre-trained.}
\label{tb:dgntable-dlg}
\end{table}

\section{Other Related Works}
\textbf{ReLU, Gating, Dual Linearity:} A spline theory based on max-affine linearity was proposed in \citep{balestriero2018spline,balestriero2018hard} to show that a DNN with ReLUs performs hierarchical, greedy template matching. In contrast, the dual view exploits the gating property to simplify the NTK into the NPK. Gated linearity was studied in \citep{sss} for single layered networks, along with a non-gradient algorithm to tune the gates. The main novelty in our work in contrast to the above is that in DLGN the feature generation is linear. We refer to the gating property of the ReLU itself and has no connection to \citep{highway} where gating is a mechanism to regulate information flow. Also, the soft-gating in our work enables gradient flow via the gating network and is different from \emph{Swish} \citep{swish}, which multiplies the pre-activation and sigmoid.\\

\textbf{Functional Role of Gating and Connection to \cite{gln}:} There are both comparable and non-comparable aspects between our work and that of Gated Linear Networks (GLNs) of \cite{gln}. The \emph{non-comparable aspects} are  (i) \text{training}: we use backpropagation to train DLGN  vs \cite{gln} propose a backpropagation free algorithm, (ii)  \text{learning in the gates:} gates are learnable in our paper vs gates are fixed and random in \cite{gln}, and (iii) \text{gating mechanism:} more importantly \cite{gln} presents only a fully connected GLN and convolutional/residual equivalents of GLNs are yet to be explored vs we have presented DLGN counterparts of state-of-the-art convolutional and residual models such as VGG-16 and ResNet-110. The \text{comparable aspects} is that the DGN/DLGN and the GLN have both separate gating and both models are essentially \emph{data dependent linear networks}. It is known that GLNs are good in continual learning tasks and given the similarity of the GLN and DLGN, an interesting future research direction would be to investigate the continual learning capabilities of the DLGN.

\textbf{Connection to \cite{gevb}.} 
The connection between our paper and that of \cite{gevb} is through the dual view. To elaborate, \cite{gevb} deal with Vectorized Non-Negative Networks (VNN) whose special case (for the case when vector dimension equals $1$) is a DNN with ReLUs with non-negative weights past first layer. As we understand, the key aspect of VNNs is the non-negativity of the weights past the first layer, and the non-negativity of the derivative of the activation function (which holds for Heaviside step function used in their paper), which enable a \emph{global error vector broadcasting} (GEVB) rule, a non-backprogagation rule. We would like to point out that in VNN, the gating is not separate (however the authors mention in the context of vectorisation unfolding over time that gating can be made separate). We use dual view to disentangle DNNs with ReLUs, whereas, \cite{gevb} use the definitions of neural path activity and neural path value to show that the GEVB rule is sign aligned with the gradient of the VNNs. In particular, it is assumed that "for all training examples, each hidden unit has at least one active path with nonzero value connecting it to the output unit" (see page 17 of \cite{gevb}). Thus, the \cite{gevb} use the dual view to justify the GEVB rule, we use the dual view to in a much more fundamental way by disentangling the computations is a DNN with ReLUs and reagrraging them in the form of DLGN to improve mathematical interepretability.

\section{Deep Gated Network : Architecture and Training}\label{sec:dgn-training}
\begin{figure}
\centering
\resizebox{0.4\columnwidth}{!}{
\begin{tikzpicture}

\node []  (fntext)at (3.5,1.5) {DGN (prior work)};
%Feature Network
\node [draw,
	minimum width=6cm,
	minimum height=1.75cm,
	thick
]  (fnbox)at (3.5,0.25) {};
\node []  (fntext)at (3.5,0.75) {Gating/Feature Network: $\Tf$};

%Feature Network Input
\node (fin) [left of=fnbox,node distance=3.5cm, coordinate] {};
\node[left=-1pt] at (fin.west){$x$};
\draw[-stealth, thick] (fin.center) -- (fnbox.west);

%Feature Network Output
\node (fout) [right of=fnbox,node distance=3.5cm, coordinate] {};
\node[right=-1pt] at (fout.west){$\hat{y}_{\text{g}}$};
\draw[-stealth, thick]  (fnbox.east)--(fout.center);

%ReLU Circle
\node[draw,
	circle,
	minimum size=0.75cm,thick,
] (relu) at (3,0){\tiny{ReLU}};
%ReLU Input
\node (b) [left of=relu,node distance=1cm, coordinate] {};

\node[left=-1pt] at (b.center){$q^\text{g}_x$};
\draw[-stealth, thick] (b.east) -- (relu.west);

%ReLU Output
\node (c) [right of=relu,node distance=1cm, coordinate] {};
\node[right=-1pt] at (c.center){$\max(0,q^\text{g}_x)$};
\draw[-stealth, thick] (relu.east) -- (c.west);

%Gating Circle
\node[draw,
	circle,
	minimum size=0.0625cm,thick,
] (gating) at (3,-1.25){\tiny{G}};
%\node[right=6pt] at (gating.north){Hard : $G(q)=\mathbbm{1}_{\{q>0\}} $};
%\node[below right=6pt] at (gating.north){Soft : $G(q)=\frac{1}{1+\exp(-\beta\cdot q)}$};

\node[right=6pt] at (3.25,-0.925){Hard : $G(q)=\mathbbm{1}_{\{q>0\}} $};
\node[right=6pt] at (3.25,-1.375){Soft : $G(q)=\frac{1}{1+\exp(-\beta\cdot q)}$};

%Value Network

\node [draw,
	minimum width=6cm,
	minimum height=1.75cm,
	thick
]  (vnbox)at (3.5,-2.625) {};

\node []  (vntext)at (3.5,-3.25) {Weight/Value Network: $\Tv$};

%Value Network Input
\node (vin) [left of=vnbox,node distance=3.5cm, coordinate] {};
\node[left=-1pt] at (vin.west){$x$};
\draw[-stealth, thick] (vin.center) -- (vnbox.west);

%Feature Network Input
\node (vout) [right of=vnbox,node distance=3.5cm, coordinate] {};
\node[right=-1pt] at (vout.west){$\hat{y}_{\text{DGN}}$};
\draw[-stealth, thick]  (vnbox.east)--(vout.center);

%GaLU Circle
\node[draw,
	circle,
	minimum size=0.75cm,thick,
] (galu) at (3,-2.5){\tiny{GaLU}};

\draw [-stealth,thick]   (b) to[out=-20,in=120] (gating.north);
\draw [-stealth,thick]   (gating.south) -- (galu.north);

%GaLU Input
\node (d) [left of=galu,node distance=1cm, coordinate] {};
\node[left=-1pt] at (d.center){$q^\text{w}_x$};
\draw[-stealth, thick] (d.east) -- (galu.west);
%GaLU Output
\node (e) [right of=galu,node distance=1cm, coordinate] {};
\node[right=-1pt] at (e.center){$q^\text{w}_x\cdot G(q^\text{g}_x)$};
\draw[-stealth, thick] (galu.east) -- (e.west);
	
\end{tikzpicture}
}
\end{figure}

The DGN is a setup to separate the gates from the weights. Consider a DNN with ReLUs with weights $\Theta\in\R^{\dnet}$. The DGN \emph{corresponding} to this DNN (left diagram in \Cref{fig:dgn}) has two networks of \emph{identical architecture} (to the DNN) namely the `gating network' and the `weight network' with distinct weights $\Tf\in\R^{\dnet}$ and $\Tv\in\R^{\dnet}$.  The `gating network' has ReLUs which turn `on/off' based on their pre-activation signals, and the `weight network' has gated linear units (GaLUs) \citep{sss,npk}, which multiply their respective pre-activation inputs by the external gating signals provided by the `gating network'.  Since both the networks have identical architecture, the ReLUs and GaLUs in the respective networks have a one-to-one correspondence.  Gating network realises $\phi_{\Tf}(x)$ by turning `on/off' the corresponding GaLUs in the weight network. The weight network realises $v_{\Tv}$ and computes the output $\hat{y}_{\text{DGN}}(x)=\ip{\phi_{\Tf}(x),v_{\Tv}}$.  The gating network is also called as the feature network since it realises the neural path features, and the weight network is also called as the value network since it realises the neural path value. 

\subsection{DGN Training}
The following two modes of training have been used in the paper namely \textbf{pretrained gates}  (PG) and \textbf{standalone training} (ST):

\emph{Pretrained Gates} (PG): The gating network is pre-trained using $\hat{y}_f$ as the output, and then the weights network is frozen, and the weights network is trained with $\hat{y}_{DGN}$ as the output. Hard gating $G(q)=\mathbbm{1}_{\{q>0\}}$ is used.

\emph{Standalone Training} (ST): Both gating and weight network are initialised at random and trained together with $\hat{y}_{DGN}$ as the output. Here, soft gating $G(q)=\frac{1}{1+\exp(-\beta\cdot q)}$ is used to allow gradient flow through gating network. We tried several values of $\beta$ in the range from $1$ to $100$, and found the range $4$ to $10$ to be suitable. We have chosen $\beta=10$ throughout the experiments.

\subsection{DLGN Training}
In all the experiments the DLGN is trained in the \textbf{standalone training} mode.

\section{Fully Connected}
Here, we present the formal definition for the neural path features and neural path values for the fully connected case in \Cref{def:formal-npf-npv}.  The layer-by-layer way of expressing the computation in a DNN of width `$w$' and depth `$d$' is given below.
\begin{table}[h]
\centering
\begin{tabular}{| ll lll|}\hline
 Input Layer &:& $z_{x,\Theta}(\cdot,0)$ &$=$ &$x$ \\
Pre-Activation&:& $q_{x,\Theta}(\iout,l)$& $=$ & $\sum_{\iin}\Theta(\iin,\iout,l) \cdot z_{x,\Theta}(\iin,l-1) $\\
Gating&:& $G_{x,\Theta}(\iout,l)$& $=$ & $\mathbf{1}_{\{q_{x,\Theta}(\iout,l)>0\}}$\\
Hidden Layer Output &:&$z_{x,\Theta}(\iout,l)$ & $=$ & $q_{x,\Theta}(\iout,l)\cdot G_{x,\Theta}(\iout,l)$\\
Final Output &:&$\hat{y}_{\Theta}(x)$ & $=$ & $\sum_{\iin}\Theta(\iin,\iout, d)\cdot z_{x,\Theta}(\iin,d-1)$\\\hline
\end{tabular}
\caption{\small{Information flow in a FC-DNN with ReLU. Here, `$q$'s are pre-activation inputs, `$z$'s are output of the hidden layers, `$G$'s are the gating values. $l\in[d-1]$ is the index of the layer, $\iout$ and $\iin$ are indices of  nodes in the current and previous layer respectively.}}
\label{tb:basic}
\end{table}

\textbf{Notation}
Index maps identify the nodes through which a path $p$ passes. The ranges of index maps $\Ifeat_l$, $\I_l,l\in[d-1]$ are $[\din]$ and $[w]$ respectively.  $\I_d(p)=1,\forall p\in[\Pfc]$.

\begin{definition}\label{def:formal-npf-npv} Let $x\in\R^{\din}$ be the input to the DNN. For this input, 

(i)  $A_{\Theta}(x,p)\eqdef\Pi_{l=1}^{d-1} G_{x,\Theta}(\I_l(p),l)$ is the activity of a path.

(ii)  $\phi_{\Theta}(x)\eqdef \left( x(\Ifeat_0(p))A_{\Theta}(x,p), p\in[\Pfc]\right)\in\R^{\Pfc}$ is the {neural path feature} (NPF).

(iii)  $v_{\Theta}\eqdef \left( \Pi_{l=1}^d \Theta(\I_{l-1}(p),\I_l(p),l), p\in[\Pfc]\right)\in\R^{\Pfc}$ is the {neural path value} (NPV).
\end{definition}
\begin{figure}[t]
\centering
\resizebox{0.9\columnwidth}{!}{
\includegraphics[scale=0.5]{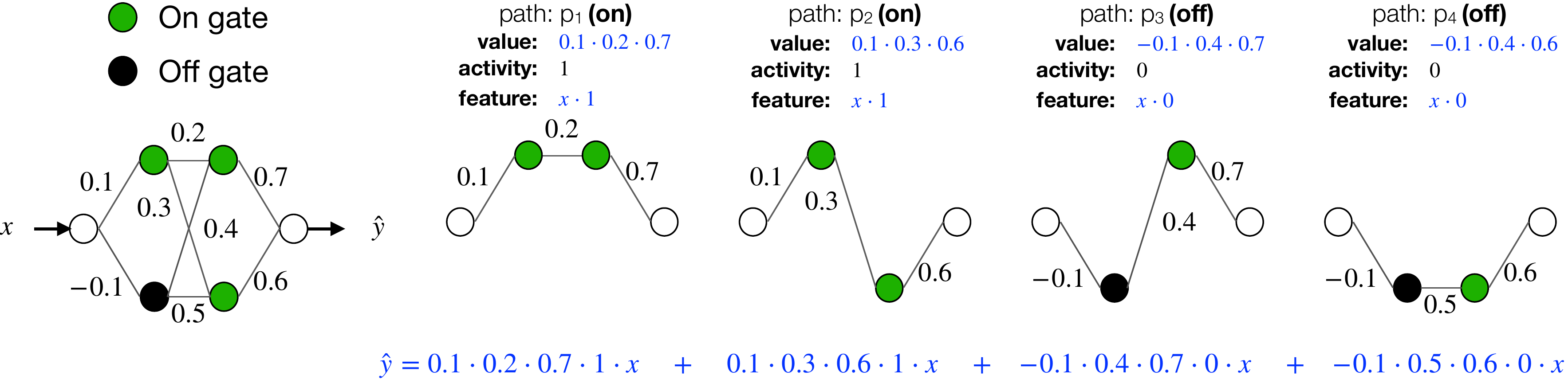}
}
\caption{Illustration of dual view in a  toy network with $2$ layers, $2$ gates per layer and $4$ paths. Paths $p_1$ and $p_2$ are `on' and paths $p_3$ and $p_4$ are `off'. The value, activity and feature of the individual paths are shown. $\hat{y}$ is the summation of the individual path contributions.}
\label{fig:paths}
\end{figure}

\section{Convolution With Global Average Pooling}\label{sec:conv}
In this section, we define NPFs and NPV in the presence of convolution with pooling. This requires three key steps (i) treating pooling layers like gates/masks (see \Cref{def:pooling}) (ii) bundling together the paths that share the same path value
(due to weight sharing in convolutions, see \Cref{def:bundle}),  and (iii) re-defining the NPF and NPV for bundles (see \Cref{def:convnps}). Weight sharing due to convolutions and pooling makes the NPK rotationally invariant \Cref{lm:cnnnpk}. We begin by describing the architecture.

\textbf{Architecture:} We consider (for sake of brevity) a $1$-dimensional\footnote{The results follow in a direct manner to any form of circular convolutions.} convolutional neural network with circular convolutions, with $\dc$ convolutional layers ($l=1,\ldots,\dc$), followed by a \emph{global-average-pooling} layer ($l=\dc+1$) and $\dfc$ ($l=\dc+2,\ldots,\dc+\dfc+1$) fully connected  layers. The convolutional window size is $\wconv<\din$, the number of filters per convolutional layer as well as the width of the FC is $w$. 

\textbf{Indexing:} Here $\iin/\iout$ are the indices (taking values in $[w]$) of the input/output filters. $\icin$ denotes the indices of the convolutional window taking values in $[\wconv]$. $\ifout$ denotes the indices (taking values in $[\din]$, the dimension of input features) of individual nodes in a given output filter. The weights of layers $l\in[\dc]$ are denoted by $\Theta(\icin,\iin,\iout,l)$ and for layers $l\in[\dfc]+\dc$ are denoted by $\Theta(\iin,\iout,l)$. The pre-activations, gating and hidden unit outputs are denoted by $q_{x,\Theta}(\ifout,\iout,l)$,  $G_{x,\Theta}(\ifout,\iout,l)$, and $z_{x,\Theta}(\ifout,\iout,l)$ for layers $l=1,\ldots, \dc$.

\begin{definition}[Circular Convolution]
For $x\in\R^{\din}$, $i\in[\din]$ and $r\in\{0,\ldots,\din-1\}$, define :

(i) $i\oplus r = i+r$, for $i+r \leq \din$ and $i\oplus r =i+r-\din$, for $i+r>\din$.

(ii) $rot(x,r)(i)=x(i\oplus r), i\in[\din]$.

(iii) $q_{x,\Theta}(\ifout,\iout,l)=\sum_{\icin,\iin}\Theta(\icin,\iin,\iout,l)\cdot z_{x,\Theta}(\ifout\oplus (\icin-1),\iin,l-1)$. 
\end{definition}
\begin{definition}[Pooling]\label{def:pooling}
Let $G^{\text{pool}}_{x,\Theta}(\ifout,\iout,\dc+1)$ denote the pooling mask, then we have
\centerline{
$z_{x,\Theta}(\iout, \dc+1) =\sum_{\ifout} z_{x,\Theta}(\ifout,\iout,\dc)\cdot G^{\text{pool}}_{x,\Theta}(\ifout,\iout,\dc+1),$
}
where in the case of \emph{global-average-pooling} $G^{\text{pool}}_{x,\Theta}(\ifout,\iout,\dc+1)=\frac{1}{\din},\forall \iout\in[w], \ifout\in[\din]$.
\end{definition}
\FloatBarrier
\begin{table}[!h]
\centering
\resizebox{1.0\columnwidth}{!}{
\begin{tabular}{|c l lll|}\hline
Input Layer&: &$z_{x,\Theta}(\cdot,1,0)$ &$=$ &$x$ \\\hline
\multicolumn{5}{l}{\quad }\\
\multicolumn{5}{l}{\quad \quad \quad \quad \quad \quad \quad \quad \quad \quad \quad \quad \quad \quad \quad \quad Convolutional Layers, $l\in[\dc]$}\\\hline
%\multicolumn{5}{l}{\quad }\\\hline
Pre-Activation&: & $q_{x,\Theta}(\ifout,\iout,l)$& $=$ & $\sum_{\icin,\iin}\Theta(\icin,\iin,\iout,l)\cdot z_{x,\Theta}(\ifout\oplus (\icin-1),\iin,l-1)$\\
Gating Values&: &$G_{x,\Theta}(\ifout,\iout,l)$& $=$ & $\mathbf{1}_{\{q_{x,\Theta}(\ifout,\iout,l)>0\}}$\\
Hidden Unit Output&: &$z_{x,\Theta}(\ifout,\iout,l)$ & $=$ & $q_{x,\Theta}(\ifout,\iout,l)\cdot G_{x,\Theta}(\ifout,\iout,l)$\\\hline
\multicolumn{5}{l}{\quad }\\
\multicolumn{5}{l}{\quad \quad \quad \quad \quad \quad \quad \quad \quad \quad \quad \quad \quad \quad \quad \quad GAP Layer, $l=\dc+1$}\\\hline
%HUO&: &${z}_{x,\Theta}(\iout,l)$ & $=$ & $\frac{1}{\din}\sum_{i\in [\din]} z_{x,\Theta}(i,\iout,l-1)$\\\hline\hline
Hidden Unit Output&: &$z_{x,\Theta}(\iout, \dc+1)$ & $=$ &$\sum_{\ifout} z_{x,\Theta}(\ifout,\iout,\dc)\cdot G^{\text{pool}}_{x,\Theta}(\ifout,\iout,\dc+1)$\\\hline
\multicolumn{5}{l}{\quad }\\
\multicolumn{5}{l}{\quad \quad \quad \quad \quad \quad \quad \quad \quad \quad \quad \quad \quad \quad \quad \quad Fully Connected Layers, $l\in[\dfc]+(\dc+1)$}\\\hline
Pre-Activation&: & $q_{x,\Theta}(\iout,l)$& $=$ & $\sum_{\iin}\Theta(\iin,\iout,l) \cdot z_{x,\Theta}(\iin,l-1) $\\
Gating Values&: &$G_{x,\Theta}(\iout,l)$& $=$ & $\mathbf{1}_{\{(q_{x,\Theta}(\iout,l))>0\}}$\\
Hidden Unit Output&: &$z_{x,\Theta}(\iout,l)$ & $=$ & $q_{x,\Theta}(\iout,l)\cdot G_{x,\Theta}(\iout,l)$\\
Final Output&: & $\hat{y}_{\Theta}(x)$ & $=$ & $\sum_{\iin}\Theta(\iin,\iout, d)\cdot z_{x,\Theta}(\iin,d-1)$\\\hline
\end{tabular}
}
\caption{Shows the information flow in the convolutional architecture described at the beginning of \Cref{sec:conv}.}
\label{tb:cconv}
\end{table}

\subsection{Neural Path Features, Neural Path Value}

\begin{proposition}
The total number of paths in a CNN is given by  $\Pcnn=\din(\wconv w)^{\dc}w^{(\dfc-1)}$.
\end{proposition}

\textbf{Notation}[Index Maps]
The ranges of index maps $\Ifeat_l$,  $\Iconv_l$, $\I_l$ are $[\din]$, $[\wconv]$ and $[w]$ respectively.

\begin{definition}[Bundle Paths of Sharing Weights]\label{def:bundle}
Let $\hat{P}^{\text{cnn}}=\frac{\Pcnn}{\din}$, and $\{B_1,\ldots, B_{\hat{P}^{\text{cnn}}}\}$ be a collection of sets such that $\forall i,j\in [\hat{P}^{\text{cnn}}], i\neq j$ we have $B_i\cap B_j=\emptyset$ and $\cup_{i=1}^{\hat{P}^{\text{cnn}}}B_i =[\Pcnn]$. Further,  if paths $p,p' \in B_i$, then $\Iconv_l(p)=\Iconv_l(p'), \forall l=1,\ldots, \dc$ and $\I_l(p)=\I_l(p'), \forall l=0,\ldots, \dc$.
\end{definition}

\begin{proposition}\label{prop:bundle}
There are exactly $\din$ paths in a bundle.
\end{proposition}

\begin{definition}\label{def:convnps} Let $x\in\R^{\din}$ be the input to the CNN. For this input, 
\begin{tabular}{rlp{12cm}}
$A_{\Theta}(x,p)$&$\eqdef$&$\left(\Pi_{l=1}^{\dc+1} G_{x,\Theta}(\Ifeat_l(p),\I_l(p),l)\right)\cdot\left(\Pi_{l=\dc+2}^{\dc+\dfc+1} G_{x,\Theta}(\I_l(p),l)\right)$\\
$\phi_{x,\Theta}(\hat{p})$&$\eqdef$&$ \sum_{\hat{p}\in B_{\hat{p}}}x(\Ifeat_0(p))A_{\Theta}(x,p)$\\
$v_{\Theta}(B_{\hat{p}})$&$\eqdef$&$ \left(\Pi_{l=1}^{\dc} \Theta(\Iconv_{l}(p),\I_{l-1}(p),\I_{l}(p),l)\right) \cdot\left( \Pi_{l=\dc+2}^{\dc+\dfc+1} \Theta(\I_{l-1}(p),\I_l(p),l)\right)$ 
\end{tabular}
\begin{center}
\begin{tabular}{|c|c|}\hline
NPF &$\phi_{x,\Theta}\eqdef (\phi_{x,\Theta}(B_{\hat{p}}),\hat{p}\in [\hat{P}^{\text{cnn}}])\in\R^{\hat{P}^{\text{cnn}}}$\\\hline
NPV& $v_{\Theta}\eqdef (v_{\Theta}(B_{\hat{p}}),\hat{p}\in [\hat{P}^{\text{cnn}}])\in\R^{\hat{P}^{\text{cnn}}}$\\\hline
\end{tabular}
\end{center}
\end{definition}

\begin{assumption}\label{assmp:main}
$\Tv_0\stackrel{\text{i.i.d}}\sim\text{Bernoulli}(\frac12)$ over $\{-{\sigma},+{\sigma}\}$ and statistically independent of $\Tf_0$.
\end{assumption}

\subsection{Rotational Invariant Kernel}
\begin{lemma}\label{lm:cnnnpk}
\begin{align*}
\text{NPK}^{\texttt{CONV}}_{\Theta}(x,x')&=\sum_{r=0}^{\din-1} \ip{x,rot(x',r)}_{\textbf{overlap}_{\Theta}(\cdot, x,rot(x',r))}\\&=\sum_{r=0}^{\din-1} \ip{rot(x,r),x'}_{\textbf{overlap}_{\Theta}(\cdot, rot(x,r),x')}
\end{align*}
\end{lemma}

\begin{proof}
For the CNN architecture considered in this paper, each bundle has exactly $\din$ number of paths, each one corresponding to a distinct input node. For a bundle $b_{\hat{p}}$, let $b_{\hat{p}}(i),i\in[\din]$ denote the path starting from input node $i$.
\begin{align*}
&\sum_{\hat{p}\in [\hat{P}]} \Bigg(\sum_{i,i'\in[\din]} x(i) x'(i') A_{\Theta}\left(x,b_{\hat{p}}(i)\right) A_{\Theta}\left(x',b_{\hat{p}}(i')\right) \Bigg)\\
=&\sum_{\hat{p}\in [\hat{P}]}\Bigg(\sum_{i\in[\din],i'=i\oplus r, r\in\{0,\ldots,\din-1\}} x(i) x'(i\oplus r) A_{\Theta}\left(x,b_{\hat{p}}(i)\right) A_{\Theta}\left(x',b_{\hat{p}}(i\oplus r)\right)\Bigg)\\
=&\sum_{\hat{p}\in [\hat{P}]}\Bigg(\sum_{i\in[\din], r\in\{0,\ldots,\din-1\}} x(i) rot(x',r)(i) A_{\Theta}\left(x,b_{\hat{p}}(i)\right) A_{\Theta}\left(rot(x',r),b_{\hat{p}}(i)\right)\Bigg)\\
=&\sum_{r=0}^{\din-1} \Bigg(\sum_{i\in[\din]} x(i) rot(x',r)(i) \sum_{\hat{p}\in [\hat{P}]}  A_{\Theta}\left(x,b_{\hat{p}}(i)\right) A_{\Theta}\left(rot(x',r),b_{\hat{p}}(i)\right)\Bigg)\\
=&\sum_{r=0}^{\din-1}\Bigg(\sum_{i\in[\din]} x(i) rot(x',r)(i) \textbf{overlap}_{\Theta}(i,x,rot(x',r))\Bigg)\\
=&\sum_{r=0}^{\din-1} \ip{x,rot(x',r)}_{\textbf{overlap}_{\Theta}(\cdot,x,rot(x',r))}
\end{align*}
\end{proof}

\begin{theorem} Let $\sigcnn=\frac{\cscale}{\sqrt{w\wconv}}$ for the convolutional layers and $\sigfc=\frac{\cscale}{\sqrt{w}}$ for FC layers. Under \Cref{assmp:main}, as $w\rightarrow\infty$, with  $\bcnn = \ \left(\dconv \sigcnn^{2(\dconv-1)}\sigfc^{2\dfc}+\dfc \sigcnn^{2\dconv}\sigfc^{2(\dfc-1)}\right)$ we have:
\begin{align*}
&\text{NTK}^{\texttt{CONV}}_{\Tdgn_0}\rightarrow\quad \frac{\bcnn}{{\din}^2} \cdot \text{NPK}^{\texttt{CONV}}_{\Tf_0}
\end{align*}
\end{theorem}

\begin{proof}
Follows from Theorem~5.1 in [\citenum{npk}].
\end{proof}
\section{Residual Networks with Skip connections}

As a consequence of the skip connections, within the ResNet architecture there are $2^b$ sub-FC networks (see \Cref{def:subfcdnn}). The total number of paths $\Pres$ in the ResNet is equal to the summation of the paths in these $2^b$ sub-FC networks (see \Cref{prop:resnetpath}). Now, The neural path features and the neural path value are $\Pres$ dimensional quantities, obtained as the concatenation of the NPFs and NPV of the $2^b$ sub-FC networks. 

\begin{proposition}\label{prop:resnetpath}
The total number of paths in the ResNet is  $\Pres = \din \cdot\sum_{i=0}^b \binom{b}{i} w^{(i+2)\dblock-1}$.
\end{proposition}

\begin{lemma}[Sum of Product Kernel]\label{lm:sumofproduct}
Let $\text{NPK}^{\texttt{RES}}_{\Theta}$ be the NPK of the ResNet, and $\text{NPK}^{\J}_{\Theta}$ be the NPK of the sub-FCNs within the ResNet obtained by ignoring those skip connections in the set $\J$. Then, \begin{align*}\text{NPK}^{\texttt{RES}}_{\Theta}=\sum_{\J\in 2^{[b]}}\text{NPK}^{\J}_{\Theta}\end{align*}
%\begin{align*}
%\end{align*}
\end{lemma}
\begin{proof}
Proof is complete by noting that the NPF of the ResNet is a concatenation of the NPFs of the $2^b$ distinct sub-FC-DNNs within the ResNet architecture.
\end{proof}

\begin{theorem} Let $\sigma=\frac{\cscale}{\sqrt{w}}$. Under \Cref{assmp:main}, as $w\rightarrow\infty$,  for $\bres^{\J} = (|\J| +2)\cdot\dblock\cdot \sigma^{2\big( (|\J|+2)\dblock-1\big)}$,
\begin{align*}
\text{NTK}^{\texttt{RES}}_{\Tdgn_0}\rightarrow \sum_{\J\in 2^{[b]}}  \bres^{\J} \text{NPK}^{\J}_{\Tf_0}
\end{align*}
\end{theorem}

\begin{proof}
Follows from Theorem~5.1 in [\citenum{npk}].
\end{proof}

\section{Numerical Experiments: Setup Details}\label{sec:expdetails}
We now list the details related to the numerical experiments which have been left out in the main body of the paper.

 $\bullet$ {Computational Resource.} The numerical experiments were run in Nvidia-RTX 2080 TI GPUs and Tesla V100 GPUs.

$\bullet$ All the models other than VGG and ResNet (and their variants) we used Adam \citep{adam} with learning rate of $3\times 10^{-4}$, and batch size of 32.

$\bullet$ All the VGG-16, Resnet-110 (and their DGN/DLGN) models  we used \emph{SGD} optimiser with momentum $0.9$ and the following learning rate schedule (as suggested in \cite{gahaalt}) : for iterations $[0, 400)$ learning rate was $0.01$,  for iterations $[400, 32000)$ the learning rate was $ 0.1$, for iterations $[32000, 48000)$ the learning rate was $0.01$, for iterations $[48000, 64000)$ the learning rate was $0.001$. The batch size was $128$. The models were trained till $32$ epochs.

\end{document}